\documentclass[10pt,twocolumn,letterpaper]{article}

\usepackage{cvpr}              

\usepackage{pifont}
\usepackage{hhline}
\usepackage[table]{xcolor}

\usepackage[most]{tcolorbox} 
\usepackage{xcolor}          

\definecolor{deepgreen}{RGB}{0,100,0}       
\definecolor{softgreen}{RGB}{80,200,120}    

\definecolor{deepblue}{RGB}{0,51,153}       
\definecolor{skyblue}{RGB}{70,130,180}      

\definecolor{deepred}{RGB}{153,0,0}         
\definecolor{softred}{RGB}{220,20,60}       

\definecolor{deeporange}{RGB}{204,85,0}     
\definecolor{softorange}{RGB}{255,140,0}    

\definecolor{deeppurple}{RGB}{102,0,153}    
\definecolor{softpurple}{RGB}{186,85,211}   

\definecolor{darkgray}{RGB}{80,80,80}       
\definecolor{lightgray}{RGB}{200,200,200}   

\definecolor{cvprblue}{rgb}{0.21,0.49,0.74}
\usepackage[pagebackref,breaklinks,colorlinks,allcolors=cvprblue]{hyperref}

\def\paperID{} 
\def\confName{CVPR}
\def\confYear{2026}

\title{RoboAgent: Chaining Basic Capabilities for Embodied Task Planning}

\author{
Peiran Xu$^{1,2}$\quad
Jiaqi Zheng$^{1}$\quad
Yadong Mu$^{1}$\thanks{Corresponding author.}\\
$^{1}$Peking University \quad $^{2}$XYZ Embodied AI\\
Beijing, China \\
{\tt\small xpr820@pku.edu.cn\quad 2400017701@stu.pku.edu.cn\quad myd@pku.edu.cn}
}

\begin{document}
\maketitle
\begin{abstract}
This paper focuses on embodied task planning, where an agent acquires visual observations from the environment and executes atomic actions to accomplish a given task. Although recent Vision-Language Models (VLMs) have achieved impressive results in multimodal understanding and reasoning, their performance remains limited when applied to embodied planning that involves multi-turn interaction, long-horizon reasoning, and extended context analysis. To bridge this gap, we propose \textbf{RoboAgent}, a capability-driven planning pipeline in which the model actively invokes different sub-capabilities. Each capability maintains its own context, and produces intermediate reasoning results or interacts with the environment according to the query given by a scheduler. This framework decomposes complex planning into a sequence of basic vision-language problems that VLMs can better address, enabling a more transparent and controllable reasoning process. The scheduler and all capabilities are implemented with a single VLM, without relying on external tools. To train this VLM, we adopt a multi-stage paradigm that consists of: (1) behavior cloning with expert plans, (2) DAgger training using trajectories collected by the model, and (3) reinforcement learning guided by an expert policy. Across these stages, we exploit the internal information of the environment simulator to construct high-quality supervision for each capability, and we further introduce augmented and synthetic data to enhance the model’s performance in more diverse scenarios. 
Extensive experiments on widely used embodied task planning benchmarks validate the effectiveness of the proposed approach. Our codes
will be available at \href{https://github.com/woyut/RoboAgent_CVPR26}{https://github.com/woyut/RoboAgent\_CVPR26}.

\end{abstract}    
\vspace{-6mm}
\section{Introduction}
\label{sec:intro}
With the rapid advancement of foundation models, the field of embodied agents has recently attracted increasing attention. To enable agents to handle complex tasks, many studies~\cite{HiRobot,pi0.5,GR1.5} adopt a hierarchical paradigm, in which the high-level planner is responsible for interpreting and decomposing task instructions, while the low-level executor generates robot control sequences. The problem of \textbf{Embodied Task Planning} (ETP)~\cite{TaPA,LLMGP,SayCan} focuses on the former, where the underlying navigation and manipulation processes are abstracted into atomic actions. The agent is required to interact with the environment by generating an appropriate sequence of atomic actions, in order to accomplish a complex task specified by the user.

\begin{figure*}[h]
  \centering
   \includegraphics[width=0.99\linewidth]{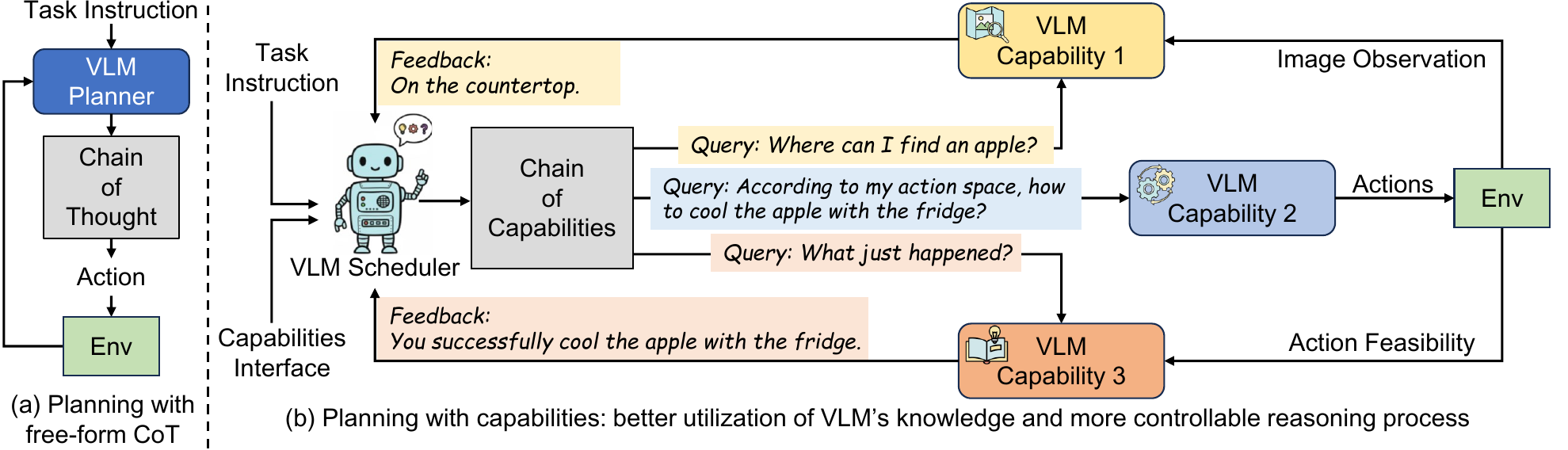}
   \vspace{-1mm}
  \caption{(a) The pipeline of a CoT-enhanced embodied task planner. (b) The pipeline of the proposed \textbf{RoboAgent} framework. By explicitly invoking specific vision–language capabilities, our method achieves a more reliable reasoning process while fully leveraging the perception and understanding proficiency of the VLM. The scheduler and the capabilities are all implemented with a single model.}
  \label{fig:motivation}
  \vspace{-5mm}
\end{figure*}

Vision-Language Models (VLMs)~\cite{Qwen2.5-VL,InterVL3.5,GLM} have shown remarkable multimodal understanding capabilities through large-scale pretraining. However, their performance in embodied planning remains suboptimal.
This gap likely stems from the intrinsic complexity of the planning problem. Unlike standard visual question answering, embodied agents must engage in multi-round interactions with the environment, perform long-horizon reasoning, and manage extensive contextual dependencies. Moreover, generating a coherent plan implicitly requires the model to accomplish multiple intermediate processes, \textit{e.g.}, intent understanding, commonsense reasoning, environment analysis, action modeling, and progress monitoring. A straightforward approach to mitigating such complexity is to decompose the planning procedure through chain-of-thought (CoT) reasoning~\cite{CoT}. Recent studies~\cite{FTLVLM,SEEA-R1,RREP} have explored methods like reinforcement learning (RL) to encourage models to produce an intermediate reasoning trace before executing an action. Although these methods have shown promising progress, the generated reasoning traces often lack principled formulation and direct supervision, making it difficult to ensure their soundness and utility for decision-making.

To bridge the gap between visual understanding and embodied planning, and to enable more reliable thought traces, we propose \textbf{RoboAgent}, a capability-driven planning framework. Specifically, we define a set of vision-language \textbf{capabilities} that are crucial for embodied scenarios. During planning, a central \textbf{scheduler} generates queries to invoke the capabilities suitable for the current context. Each capability functions as an additional layer between the planner and the environment. It either produces intermediate reasoning results or generates atomic actions for interaction.
This framework offers several benefits. (1) It effectively leverages the inherent competencies of the underlying VLM to simplify the overall planning process. (2) It yields a more controllable and transparent reasoning process. During training, this allows us to apply fine-grained supervision for the intermediate thoughts; during inference, it facilitates clear diagnosis of failure cases and performance bottlenecks. (3) Unlike the works with tool-augmented language model~\cite{Toolformer,Hugginggpt,VisProg}, the scheduler and all capabilities in our methods are implemented with a single, end-to-end trainable VLM, eliminating the need for external dependencies. This design reflects our belief that modern VLMs are inherently capable of handling all aspects of embodied reasoning, and what is required is an appropriate mechanism to invoke their abilities.

We propose a multi-stage training strategy for effectively fine-tuning the VLM. We begin with a standard practice of supervised fine-tuning (SFT) on expert data. Beyond conventional expert action trajectories, we further leverage internal information from the environment simulator (\textit{e.g.}, object locations, segmentation masks, action feedbacks) to construct dedicated training datasets for each capability. This privileged information is inaccessible to the agent during inference, while it provides high-quality supervision for the reasoning process during training.
Subsequently, we deploy the trained model in the environment to collect new trajectories with capability invocation sequences, and construct corrective ground-truth annotations for the involved capabilities, enabling a DAgger-style~\cite{DAgger} training procedure. For the scheduler, we further develop an expert-guided policy optimization algorithm for reinforcement fine-tuning (RFT), and introduce synthetic interaction data to enlarge the training set.
Together, these stages progressively enhance the model’s performance on challenging tasks and its generalization to novel scenarios.

The contributions of this work can be summarized as follows:
\begin{itemize}
    \item We formulate a capability-driven embodied planning pipeline that decomposes a complex planning task into a series of simpler vision-language problems.
    \item We propose a multi-stage training pipeline to optimize the VLM for planning, leveraging intermediate supervision and diverse sources of data.
    \item We conduct experiments on multiple simulated environments and benchmarks to validate the effectiveness and generalizability of the proposed method.
\end{itemize}

\section{Related Works}
\subsection{Embodied Task Planning}
Task and Motion Planning~\cite{PRT,PRMPP,PDDL,PDDL2.1,HTAMP,CTAMP} is a classic problem in robotics. In recent years, the commonsense knowledge embedded in large language models (LLMs) has enabled planning in more open-ended environments and for more diverse tasks. Early efforts on LLM-based embodied planning~\cite{LMZSP,LMIDM} are primarily conducted in a text world, focusing on research directions such as plan representation~\cite{ProgPrompt,FLTRNN}, world modeling~\cite{LLMCK,WorMI}, error correction and self-refining~\cite{CRP,CAPE,ESM}, decoding strategies~\cite{TreePlanner}, and training data construction~\cite{LMWM}.

With the advancement of vision foundation models, a more realistic setting of planning based on visual observations has become increasingly prevalent. One common approach is to leverage off-the-shelf closed-source models. Some works have explored enhancements such as incorporating scene graphs as environment representations~\cite{LookLeap,ESCA}, introducing multi-agent cooperation frameworks~\cite{CoELA,CaPo,CoBel-World,REVECA,LIET}, designing memory modules~\cite{RoboMemory,EmbodiedRAG,EPoT} and replanning strategies~\cite{CLEA,ExploreVLM}, and extending the scale and complexity of the tasks~\cite{BUMBLE,EWA,ExRAP}. Another line of works, including this paper, considers the training of open-source models. \cite{LightPlanner,Embodied-Reasoner,WAP} perform SFT using expert trajectories enhanced with CoT. \cite{LLMGP,GMLLMA,Octopus,VIPER,FTLVLM,GTR,SEEA-R1,RREP,RLVMR,ERA,RLSW,REPVR} employ RL algorithms~\cite{PPO,Deepseekmath,DAPO} with carefully designed reward functions. \cite{D2PO,EMMA,MCTS-EP,TCPO,SPO} collect offline data through methods such as tree search, and train models with direct preference optimization~\cite{DPO}.
In addition, there are also studies focusing on developing generalist foundation models for embodied intelligence~\cite{EmbodiedGPT,RoboBrain,RoboBrain2,GEA,EmbodiedBrain,Robix,GR1.5,Vlaser,BEAR,Cosmos-Reason1}, as well as creating more comprehensive evaluation benchmarks~\cite{TaPA,EMBODIEDBENCH,VAB,InfThor,MuEP,EmbodiedArena,EmbodiedEval,RoboBench,VLMZS,CookBench,PARTNR,VIKI-R}.

\subsection{Reasoning with Large Models}
Reasoning has become a central topic in the era of large models. In the context of embodied agents, a growing body of works explores how guiding models to generate intermediate reasoning steps can improve the effectiveness of their actions~\cite{ReAct,Reflexion}. From an architectural perspective, \cite{Plan-and-Act,MPO,HiPlan,PilotRL,InstructFlow,ReAcTree} investigate progressive reasoning pipelines, in which tasks are first decomposed into sub-tasks or sub-goals before concrete actions are generated. \cite{ARMAP,LAC,SLQ} discuss guiding the decoding process of LLMs with a learned reward model or Q model. From the perspective of optimization, \cite{GiGPO,UETPA,DynaMind,EMPG,Search-R1} enhance existing RL algorithms to better handle multi-turn interactions, while \cite{IPR,ADAPT,AgentQ} propose new preference optimization frameworks to align model behaviors. In terms of data, \cite{AgentBank,AgentGym,DeepAgent} expand the scale and diversity of training datasets, while \cite{BPO,Agent-R,FTRAG,TnE} focus on leveraging self-generated trajectories for iterative self-improvement.

This work aims to achieve a more controllable and reliable reasoning process through the explicit invocation of capabilities. Unlike existing progressive planning frameworks~\cite{Plan-and-Act,MPO,InstructFlow} that bridge the task and actions via sub-tasks, we design the intermediate layer as specific capabilities, thereby leveraging the vision-language knowledge inherently embedded in the VLM. In addition, unlike existing reasoning models that tackle complex problems through self-questioning~\cite{Socratic,L2M,LM2,DRC,SQLM}, our method introduces principled capability interfaces, which facilitate fine-grained supervision over the reasoning steps.
Finally, in contrast to prior methods that rely on closed-source models or external tools~\cite{WT,ExploreVLM,LightPlanner}, our approach implements all modules within a single open-source VLM, complemented by a carefully designed multi-stage training strategy.

\section{Method}
\label{sec:method}
\begin{figure*}[h]
  \centering
  \includegraphics[width=0.97\linewidth]{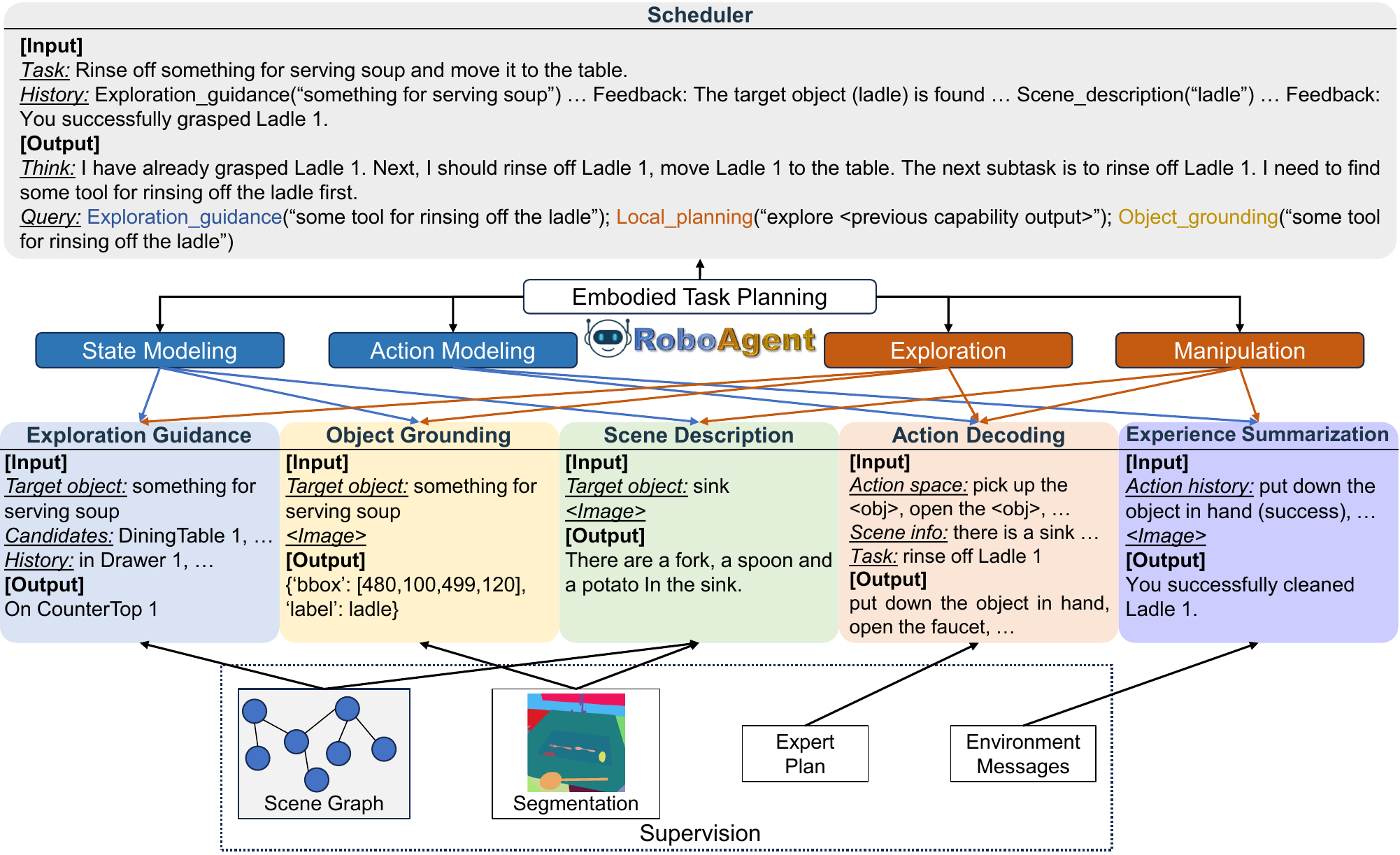}
  \vspace{-2mm}
  \caption{An illustration of the scheduler and 5 capabilities involved in our RoboAgent.}\label{fig:method}
  \vspace{-5mm}
\end{figure*}

\subsection{Formulation}
In the problem of ETP, an agent is required to complete a task by performing a sequence of atomic actions in the environment. Specifically, the agent receives a text instruction $I$ at the beginning of an episode. Then for each timestep $t$, it gets an egocentric RGB image $o_t$ as observation, and outputs an action $a_t$ chosen from a pre-defined action set $\mathcal{A}$. The environment state will evolve in response to the action of the agent: $s_{t+1}=\mathcal{E}(s_t, a_t)$, where $\mathcal{E}$ is implemented by a simulator. At the end of an episode, evaluation is performed by checking whether the final environment state satisfies the goal conditions $\left\{\mathbf{r}_i\right\}_{i=1}^{N_{\text{goal}}}$, where $\mathbf{r}_i(s_T)=1/0$ is a check on object states or relations.

\subsection{Capability-Driven Planning}
As described in Sec.~\ref{sec:intro} and Fig.~\ref{fig:motivation}, RoboAgent implements a scheduler and multiple capabilities with a single VLM to perform capability-driven planning. Formally, 
\begingroup
\setlength{\abovedisplayskip}{3pt}   
\setlength{\belowdisplayskip}{0pt}   
\begin{equation}\label{eq:scheduler}
    \mathbf{M}(I,p^S,c^S_i)=\left[(g_i^j, q_i^j)\right]_{j=1}^{n_i},
\end{equation}
\endgroup
\begingroup
\setlength{\abovedisplayskip}{0pt}   
\setlength{\belowdisplayskip}{1pt}   
\begin{equation}
    \mathbf{M}(p^{g_i^j},q_i^j,o_i^j)=(\textbf{a}_i^j,f_i^j),
\end{equation}
\endgroup
where $\mathbf{M}$ is the VLM, $[g_i^j]^{n_i}_{j=1}$ are $n_i$ capabilities invoked sequentially at the $i$-th scheduler calling, $p^S$ and $p^{g_i^j}$ are the prompts for the scheduler and the capability, $c^S_i$ is the context maintained by the scheduler, $q_i^j$ is the query for capability $g_i^j$, $o_i^j$ is the (optional) image input for capability $g_i^j$, $\textbf{a}_i^j$ is a sequence of generated actions, $f_i^j$ is the generated text serving as the feedback from the capability to the scheduler. $q_i^j$ and $f_i^j$ are subsequently integrated with $c^S_i$ to produce the context for the next scheduler calling. 

We implement 5 capabilities for RoboAgent in this work: $g_i^j\in\mathcal{C},\mathcal{C}=\{$EG, OG, SD, AD, ES$\}$. Specifically, \textbf{Exploration Guidance} (EG) takes a target object as input and, based on the commonsense knowledge of scene layouts and object placements, predicts the most promising exploration direction in order to find the object. \textbf{Object Grounding} (OG) performs open-vocabulary grounding to determine whether a specific object is currently observable within the agent’s field of view. \textbf{Scene Description} (SD) produces a textual description of the state of a target object, forming the basis for subsequent manipulation. \textbf{Action Decoding} (AD) translates a navigation or manipulation command into an executable sequence of atomic actions within $\mathcal{A}$. \textbf{Experience Summarization} (ES) summarizes the interaction outcome of the most recent action sequence generated by AD, and analyzes the cause of failure when errors occur.
Among the five capabilities, AD generates a sequence of actions without additional outputs ($\textbf{a}_i\neq\emptyset, f_i=\emptyset$), while the other four capabilities do not generate actions but instead provide textual feedback to the scheduler ($\textbf{a}_i=\emptyset, f_i\neq\emptyset$). Fig.~\ref{fig:method} illustrates the functionality of each capability, while the detailed input–output formats and prompts design are provided in the Appendix.

\vspace{-1mm}
\subsection{Stage 1: Training with Expert Trajectories}\label{subsec:SFT}

To make the VLM compatible with our planning pipeline, we first perform SFT on the expert trajectories. Given a task and its expert plan, we construct \textbf{training samples for the scheduler} as follows. We begin by identifying the key objects involved in the task, \textit{i.e.}, the objects directly appearing in the goal conditions and the tools required to achieve the conditions (\textit{e.g.}, a \texttt{Fridge} is required for making an \texttt{apple} \texttt{cool}).
Next, using the target objects as anchors, the expert trajectory is segmented into a sequence of exploration and manipulation sub-plans. Each exploration sub-plan searches several candidate regions and ends when a new target object is found, while each manipulation sub-plan performs consecutive control actions on the target object to alter its position or state.
For every exploration sub-plan, the scheduler must iteratively examine possible locations for the target object until it enters the field of view. We map this process into EG, AD, and OG capabilities, where EG proposes an exploration direction, AD converts the exploration command into navigational actions, and OG determines whether the object becomes observable.
For each manipulation sub-plan, the scheduler is required to analyze the current state of the target object and select proper actions to accomplish the intended change. This process is represented as a sequence of SD, AD, and ES capabilities, where SD describes the key information in the scene, AD generates control actions accordingly, and ES summarizes the execution outcomes. In this way, the expert plan is transformed into a sequence of capability invocations.

To generate the query for each capability calling, we feed the task instruction and goal conditions into an off-the-shelf LLM, which parses the instruction into a set of object descriptions (\textit{e.g.}, the instruction ``Rinse off something for serving soup and move it to the table" in Fig.~\ref{fig:method} is parsed into ``something for serving soup", ``some tool for rinsing", and ``table"). These descriptions are then used as queries for the capabilities involved in exploration sub-plans.
For manipulation sub-plans, we employ a rule-based approach to categorize them into predefined types such as ``grasp something", ``place something somewhere", or ``clean something with some tool", and then substitute the placeholders with the parsed object descriptions to form the required queries.

In addition, to enhance the reasoning ability of the scheduler, we construct CoT traces using a template-based method. Each trace contains a retelling of the task instruction, a list of all sub-plans, a list of completed sub-plans, and the target of the next sub-plan.
The resulting combination of thinking process, capabilities, and queries forms the ground-truth output for each scheduler calling (Eq (\ref{eq:scheduler})).

On the other hand, to create the \textbf{training samples for the capabilities}, we fully leverage the internal information of the environment simulator as supervision signals. Specifically, we extract the following items from the simulator while executing the expert plan: (1) the scene graph (SG), which captures the properties and relations of the objects within the environment; (2) observation images along with their instance segmentation masks; (3) environment messages, which indicate whether each action succeeds and, if not, the reason for failure. 
The object location information in the SG provides ground-truth answers for EG. Segmentation masks of the target object are converted into JSON-formatted bounding box annotations, serving as ground-truth for OG. By filtering the SG with object IDs appearing in the segmentation masks, we obtain the sub-graph corresponding to the agent’s partial observation at each timestep. Extracting the states and relations relevant to the target object from this filtered graph and converting them into text yields ground-truth for SD. Environment messages are used as ground-truth for ES. Finally, supervision for AD comes directly from the corresponding action sequences in the expert trajectories.

\subsection{Stage 2: Training with Model-Generated Data}\label{subsec:dagger}

Fine-tuning on expert trajectories improves the model’s basic abilities but constrains its exploration behavior and adaptability to unseen states beyond the training distribution. To address this limitation, we apply the fine-tuned model to the training tasks to collect model-generated plans (whether successful or failed) and the corresponding chains of capabilities.
For SD and ES invocations, we adopt the same strategy as in Sec.~\ref{subsec:SFT} to construct supervision from the SG and environment messages.
For EG and OG, we measure the semantic similarity between each query issued by the model and the queries in the ground-truth invocation sequence. If the highest similarity exceeds a certain threshold, indicating a meaningful invocation, we retrieve the corresponding target object from the matched ground-truth query and generate supervision for that object using the SG and segmentation masks.
For AD, we examine whether its query belongs to one of the predefined sub-plan categories, and construct the ground-truth action sequence based on the target object's state recorded in the SG.
Through this process, we assign corrective supervision to each capability invocation within the model-generated trajectories. Mixing these samples with the expert dataset yields a DAgger-style~\cite{DAgger} training procedure for the capabilities.

Furthermore, we perform data augmentation on the training set to enhance its diversity. On one hand, we employ an LLM to generate multiple descriptive phrases for each target object category. These phrases are used to replace the original object references in capability queries, improving the open-vocabulary comprehension of OG and EG.
On the other hand, we modify the format of the atomic actions (\textit{e.g.}, replacing ``pick up the object” with ``grasp the object'') to strengthen the robustness of AD and ES under different action spaces, allowing them to acquire more generalized action knowledge.

\subsection{Stage 3: Training with Expert Policy}\label{subsec:RFT}
In training stage 2, we expanded the capability training to a larger data scope by leveraging annotations derived from the simulator. However, constructing supervision for the scheduler on newly collected data is more challenging, especially when the output involves a CoT process.
To further enhance the reasoning ability of the scheduler, we introduce an additional RFT stage in which the model receives rewards for making appropriate capability invocations.
In what follows, we first present the derivation of the algorithm used in this stage, then describe its practical implementation.

Let $\eta(\pi)$ be the expected return of a policy $\pi$, \textit{i.e.}, $\eta(\pi)=\mathbb{E}_{(s_0,a_0,...,s_T,a_T)\sim\pi}[\sum_{t=0}^T\gamma^tR(s_t,a_t)]$\footnote{We consider a general policy in the derivation. In the context of our scheduler, $s_t$ denotes its input at the $t$-th step, while $a_t$ represents the invoked capabilities and corresponding queries.}.
From~\cite{AOARL,TRPO} we have: for any policies $\pi$, $\pi'$,
\begingroup
\setlength{\abovedisplayskip}{0pt}   
\setlength{\belowdisplayskip}{2pt}   
\begin{equation}
    \eta(\pi)=\eta(\pi')+\mathbb{E}_{(s_0,a_0,...,s_T,a_T)\sim\pi}\sum_{t=0}^T\gamma^tA_{\pi'}(s_t,a_t).
    \label{eq:return_diff}
\end{equation}
\endgroup
Assuming the access to an expert policy $\pi^*$ that is able to successfully complete all the tasks, we replace $\pi'$ with $\pi^*$ in Eq (\ref{eq:return_diff}) and obtain a straightforward target for optimizing the policy $\pi$:
\begingroup
\setlength{\abovedisplayskip}{3pt}   
\setlength{\belowdisplayskip}{3pt}   
\vspace{-3mm}
\begin{align}
    J(\pi) &= \mathbb{E}_{(s_0,a_0,\dots,s_T,a_T)\sim\pi} \sum_{t=0}^T \gamma^t A_{\pi^*}(s_t,a_t) \\
           &= (1-\gamma) \mathbb{E}_{s\sim d_{\pi}} \mathbb{E}_{a\sim \pi(\cdot|s)} A_{\pi^*}(s,a).
    \label{eq:J}
\end{align}
\endgroup
Compared with TRPO and PPO that optimize the gain in expected return ($\eta(\pi')-\eta(\pi)$), Eq (\ref{eq:J}) aims to maximize the expert's advantage. During batch training, the advantage $A_{\pi^*}$ can be accurately computed with the deterministic expert policy, thus avoiding the variance introduced by Monte Carlo estimation of future returns.

Due to the high cost of collecting interaction trajectories for Eq (\ref{eq:J}) during policy training, we build an offline dataset of states, $D$, which is similar to the prompt dataset for classic RLHF. We also introduce importance sampling (IS) to enable training with off-policy action samples:
\begingroup
\setlength{\abovedisplayskip}{3pt}   
\setlength{\belowdisplayskip}{3pt}   
\begin{equation}
    J(\pi)=\mathbb{E}_{s\sim D} \mathbb{E}_{a\sim \pi_{\text{old}}(\cdot|s)}[\frac{\pi(a|s)}{\pi_{\text{old}}(a|s)}A_{\pi^*}(s,a)].
    \label{eq:J-IS}
\end{equation}
\endgroup
Following PPO~\cite{PPO}, the probability ratio $r(a,s)=\frac{\pi(a|s)}{\pi_{\text{old}}(a|s)}$ will be clipped to avoid large policy deviation.
Further, we insert a GRPO-style~\cite{Deepseekmath} group-based average into $J$ as a baseline:
\begingroup
\setlength{\abovedisplayskip}{1pt}   
\setlength{\belowdisplayskip}{1pt}   
\begin{equation}
    J(\pi)=\mathbb{E}_{s\sim D}\mathbb{E}_{a^i\sim\pi_{\text{old}}(\cdot|s)}\frac{1}{G}\sum_{i=1}^G[r(a^i,s)\hat{A}_{\pi^*}(s,a^i)],
    \label{eq:J-baseline}
\end{equation}
\endgroup
\begingroup
\setlength{\abovedisplayskip}{1pt}   
\setlength{\belowdisplayskip}{0pt}   
\begin{equation}
    \hat{A}_{\pi^*}(s,a^i)=A_{\pi^*}(s,a^i)-\frac{1}{G}\sum_{j=1}^GA_{\pi^*}(s,a^j),
    \label{eq:J-baseline-A}
\end{equation}
\endgroup
where the subtracted average works as an estimate of $\mathbb{E}_{a\sim\pi_{\text{old}}(\cdot|s)}[A_{\pi^*}(s,a)]$ to reduce the variance of policy gradient.
Also, due to the optimality of $\pi^*$, it can be shown that $A_{\pi^*}(s,a)\le0$ for all $(s,a)$, and the equality only holds when $a$ is an optimal action at state $s$. Therefore, when optimizing Eq (\ref{eq:J-IS}), all suboptimal actions are suppressed, while the optimal actions receive zero gradients. Although this mechanism will drive the policy towards selecting optimal actions, the absence of ``positive" signals may lead to slow convergence. By introducing a baseline as Eq (\ref{eq:J-baseline}), relatively better (though not necessarily optimal) actions within the group are encouraged, while worse actions remain suppressed, resulting in a more gradual and progressive learning process.

\begin{figure}[t]
  \centering
  \includegraphics[width=0.95\linewidth]{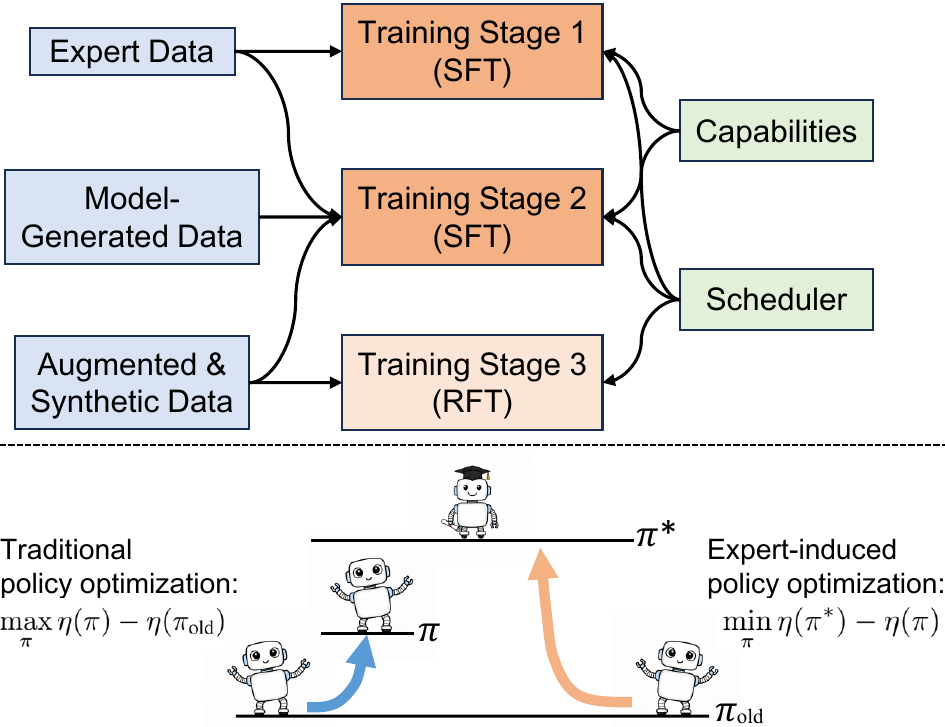}
  \vspace{-3mm}
  \caption{Upper: illustration of the data types and modules involved in different training stages.
Lower: demonstration of the training objective in the RFT stage. Traditional RL algorithms optimize the improvement of return over the original policy, whereas the proposed EIPO optimizes the discrepancy between the policy return and the expert return.}\label{fig:train}
\vspace{-7mm}
\end{figure}

We refer to the algorithm of optimizing Eq (\ref{eq:J-baseline}) as \textbf{E}xpert-\textbf{I}nduced \textbf{P}olicy \textbf{O}ptimization (EIPO). Its framework is similar to that of PPO and GRPO. A key difference lies in its use of a more stable objective function, $A_{\pi^*}$. Unlike $A_{\pi}$ that requires rollouts of the learned policy $\pi$ for estimation, $A_{\pi^*}$ can be directly computed from the expert policy $\pi$. Now, in order to employ EIPO to train the scheduler, we need to formulate an expert scheduler $\pi_{S}^*$, a reward function $R$ for advantage computation, and a dataset $D$. To implement the expert, we monitor the completion status of each sub-plan and sequentially convert the unfinished sub-plans into corresponding capability invocations (Sec.~\ref{subsec:SFT}). Empirically, this expert policy can accomplish the task from most states, assuming that the capabilities produce correct outputs.
We assign a reward of +1 to the scheduler when its invoked capability completes a manipulation sub-plan. Thus, leveraging the list of unfinished sub-plans, we can compute the value function of $\pi^*$ for any given state and thereby compute $A_{\pi^*}$.
Regarding the training samples, we note that the scheduler only requires textual feedback from the Capabilities during execution, without the need for direct interaction with the environment. Given a task, we employ the expert scheduler to generate capability invocations. For each capability, feedback messages are synthesized with a certain probability of error, forming the multi-turn interaction context that serves as the model input for EIPO training.
During data construction, we apply the augmentations for object description and action-space introduced in Sec.~\ref{subsec:dagger}, and further enhance task-level diversity by merging multiple training tasks into composite ones. This enables the scheduler to generalize to more diverse and challenging task scenarios.

\begin{table*}
  \caption{Performance comparison on EB-ALFRED~\cite{EMBODIEDBENCH}.}
  \vspace{-2mm}
  \label{tab:main-EBALF}
  \centering
  \scalebox{0.86}{
  \begin{tabular}{lc>{\columncolor{blue!10}}ccccccc}
    \toprule
      & Base Model & \textbf{Avg} & Base & Common & Complex & Visual & Spatial &  Long \\
    \midrule
  Zero-Shot\cite{EMBODIEDBENCH} & GPT-4o & 56.3 & 64 & 54 & 68 & 46 & 52 & 54 \\ 
  & Claude-3.7-Sonnet & \textbf{67.7} & 68 & \textbf{68} & 70 & \textbf{68} & \textbf{62} & \textbf{70} \\  
  & Gemini-1.5-Pro & 62.3 & \textbf{70} & 64 & \textbf{72} & 58 & 52 & 58 \\ 
  & Qwen-VL-Max & 41.3 & 44 & 48 & 44 & 42 & 38 & 32 \\ 
  & Qwen2.5-VL-72B & 39.7  & 50  & 42  & 42  & 36  & 34  & 34 \\ 
  \midrule
  RoboBrain2.0~\cite{RoboBrain2} & Qwen2.5-VL-7B & 14.0 & -  & -  & -  & -  & -  & - \\ 
  Vlaser~\cite{Vlaser} & InternVL3-8B & 50.0 & -  & -  & -  & -  & -  & - \\ 
  \midrule
  REBP~\cite{RREP} & Qwen2.5-VL-7B & 35.6 & 54  & 42  & 46  & 28  & 38  & 6 \\
  WAP~\cite{WAP} & Qwen2.5-VL-7B & 62.7 & 66 & \textbf{62} & \textbf{70} & 56 & 52 & 70 \\
  RoboGPT-R1~\cite{RoboGPT-R1} & Qwen2.5-VL-3B & 55.3 & 62 & 56 & 64 & 50 & 50 & 50 \\
  \midrule
  \textbf{RoboAgent} (Ours) & Qwen2.5-VL-3B & \textbf{67.0} & \textbf{72} & 48 & 64 & \textbf{78} & \textbf{60} & \textbf{80} \\
  \bottomrule
  \end{tabular}
  }
  \vspace{-1mm}
\end{table*}

\begin{table*}[t]
\centering
\begin{minipage}{0.65\textwidth}
\caption{Performance comparison on ALFWorld~\cite{ALFWorld} (visual observation).}
\vspace{-1mm}
  \label{tab:main-AW}
  \centering
\scalebox{0.79}{
  \begin{tabular}{lc>{\columncolor{orange!20}}ccccccc}
    \toprule
   & Base Model & \textbf{Avg} & Pick & Clean & Heat & Cool & Look & Pick2 \\
    \midrule
  Zero-Shot & GPT-4V & 19.4 & 38 & 18 & 6.7 & 18 & 12 & 15 \\
  (reported in  & Gemini & 13.5 & 35 & 0 & 0 & 0 &16 & 12 \\
 \cite{GTR,SEEA-R1}) & GPT-4o & 24.0 & 44 & 22 & 29 & 27 & 7 & 23 \\
  \midrule
  GTR~\cite{GTR} & LLaVA1.6-mistral-7b & 17.0 & 37 & 7 & 8 & 33 & 23 & 20 \\
  RL4VLM~\cite{FTLVLM} & LLaVA1.6-mistral-7b & 21.7 & 47 & 10 & 14 & 19 & 15 & 18 \\
  GFlowVLM~\cite{GFlowVLM} & LLaVA1.6-mistral-7b & 26.1 & 50 & 10 & 19 & 24 & 23 & 24 \\
  TCPO~\cite{TCPO} & LLaVA1.6-mistral-7b & 26.7 & 27 & 25 & 29 & 6 & 33 & 42 \\
  CoSo~\cite{CoSo} & LLaVA1.6-mistral-7b & 26.5 & 42 & 21 & 12 & 22 & 21 & 26 \\
  SEEA-R1~\cite{SEEA-R1} & Qwen2.5-VL-7B & 36.0 & 49 & 40 & 43 & 41 & 24 & 16 \\ 
  
  \midrule
  \textbf{RoboAgent} (Ours) & Qwen2.5-VL-3B & \textbf{77.6} & \textbf{92} & \textbf{84} & \textbf{74} & \textbf{57} & \textbf{94} & \textbf{59} \\
  \bottomrule
  \end{tabular}
}
\end{minipage}
\hfill
\begin{minipage}{0.33\textwidth}
\centering
\caption{Performance comparison on ALFWorld~\cite{ALFWorld} (textual observation).}
\vspace{-1mm}
  \label{tab:main-AW-text}
  \centering
  \scalebox{0.72}{
  \begin{tabular}{lccc}
    \toprule
  & Base Model & seen & unseen \\
    \midrule
 Zero-Shot~\cite{SEEA-R1}  & GPT-4o & 78.6 & 83.6 \\
\midrule
ETO~\cite{TnE} & Llama2-7B & 68.6 & 72.4 \\
IPR~\cite{IPR} & Llama-2-7B & 70.3 & 74.7 \\
MPO~\cite{MPO} &Llama3.1-8B & 85.0 & 79.1 \\
LAC~\cite{LAC} & Gemma-7B & - & 84.0 \\
SEEA-R1~\cite{SEEA-R1} & Qwen2.5-7B & 85.3 & 85.1 \\
BPO~\cite{BPO} & Llama3.1-8B & 87.9 & 89.6 \\
GiGPO~\cite{GiGPO} & Qwen2.5-7B & 90.8 & - \\ 
DynaMind~\cite{DynaMind} &  Qwen2.5-7B & \textbf{92.5} & 89.1 \\
\midrule
\textbf{RoboAgent} (Ours) & Qwen2.5-VL-3B & 92.1 & \textbf{94.0}\\
  \bottomrule
  \end{tabular}
  }
\end{minipage}
\hfill
\vspace{-5mm}
\end{table*}

\section{Experiments}
\subsection{Training Dataset}\label{subsec:data}
ALFRED~\cite{Alfred} is originally proposed as an embodied instruction following simulator~\cite{FILM,CAPEAM,EIFUE}. In recent years, many studies~\cite{ALFWorld,LoTa-Bench,EMBODIEDBENCH,MuEP,EmbodiedBrain} have adapted it into an ETP environment by encapsulating atomic actions. The dataset is divided into training, validation, and test splits, where the training split contains 6,374 tasks accompanied by 20k human-annotated instructions. Using these training tasks and the action spaces defined by~\cite{ALFWorld} and~\cite{EMBODIEDBENCH}, we follow the procedure described in Sec.~\ref{subsec:SFT} to generate a total of 640k training samples for the capabilities and the scheduler. They are used for the first SFT training stage.
Subsequently, we collect the trajectories generated by the model on the training tasks and apply the data augmentation introduced in Sec.~\ref{subsec:dagger}, resulting in 690k samples for the second DAgger-style training stage. Finally, we synthesize 25k trajectories and samples from them to form the training set for the RFT stage. Further details regarding the format and statistics of the training data can be found in the Appendix.

\begin{table*}[t]
\centering
\label{tab:three-tables}

\begin{minipage}{0.48\textwidth}
\caption{Analysis on the effect of different training stages and different sources of training data.}
  \label{tab:main-abl}
  \centering
  \begin{tabular}{lccc>{\columncolor{orange!20}}c>{\columncolor{blue!10}}c}
    \toprule
    & SFT-expert & SFT-dagger  &  RFT & AW & EB \\
    \midrule
  (1) & \ding{51} & \ding{55} & \ding{55} & 44.8 & 62.0\\
  (2) & \ding{51} & aug. exp. & \ding{55} & 41.8 & 64.7\\
  (3) & \ding{51} & aug. gen.& \ding{55} & 73.1 & 64.3\\
  (4) & \ding{51} & aug. gen. & exp. & 71.6 & 64.7 \\
  (5) & \ding{51} & aug. gen. & aug. exp. & 74.6 & 65.7\\
  (6) & \ding{51} & aug. gen. & aug. syn. & \textbf{77.6} & \textbf{67.0}\\
  \bottomrule
  \end{tabular}
\end{minipage}
\hfill
\begin{minipage}{0.24\textwidth}
\centering
\small
\caption{OOD results on EB-Habitat~\cite{EMBODIEDBENCH}.}
  \label{tab:ood-hab}
  \centering
  \begin{tabular}{lc}
    \toprule
    & SR \\
    \midrule
    GPT-4o & \textbf{59.0} \\
    Claude-3.7-Sonnet & 58.7 \\
    Gemini-1.5-Pro & 56.3 \\
\midrule
REBP~\cite{RREP} & 20.0 \\ 
RoboGPT-R1~\cite{RoboGPT-R1} & 22.0 \\
\textbf{RoboAgent} (Ours) & \textbf{22.3} \\
  \bottomrule
  \end{tabular}
\end{minipage}
\hfill
\begin{minipage}{0.22\textwidth}
\centering
\small
\caption{OOD results on LoTa-WAH~\cite{LoTa-Bench}, using subgoal success rate as metric.}
  \label{tab:ood-lota}
  \centering
  \begin{tabular}{lc}
    \toprule
    & SSR \\
    \midrule
    GPT-3.5-turbo & 36.0 \\
    GPT-4 & 37.4 \\
\midrule
LLaMA-7B~\cite{LoTa-Bench} & 3.7 \\
LLaMA-30B~\cite{LoTa-Bench} & 10.4 \\
\textbf{RoboAgent} (Ours) & \textbf{22.1}\\
  \bottomrule
  \end{tabular}
\end{minipage}
\vspace{-5mm}
\end{table*}

\subsection{Evaluation Benchmarks}
We evaluate the proposed method in two commonly used ETP environments, ALFWorld~\cite{ALFWorld} and EB-ALFRED~\cite{EMBODIEDBENCH}. Both benchmarks are built upon the AI2-THOR simulator~\cite{AI2-THOR} and ALFRED dataset~\cite{Alfred}, but differ in their action spaces and instruction styles. It is worth noting that our training process relies solely on the training split of ALFRED~\cite{Alfred}, without using the test tasks from ALFWorld or EB (which are derived from ALFRED’s validation set). Consequently, our evaluation imposes strict requirements on the model’s generalization to unseen scenes and novel instructions. To validate the model's performance on a wider range of task domains, we also conduct out-of-domain (OOD) evaluation on EB-Habitat~\cite{EMBODIEDBENCH,LLMGP} and LoTa-WAH~\cite{LoTa-Bench,WAH}, which are built on the Habitat~\cite{Habitat2} and VirtualHome~\cite{VirtualHome} simulators, respectively. We utilize success rate (SR) as the metric for ALFWorld and EB, and subgoal success rate (SSR) for LoTa.

\subsection{Implementation Details}
We employ Qwen2.5-VL-3B~\cite{Qwen2.5-VL} as the base VLM. As described in Sec.~\ref{sec:method}, it first goes through an expert-SFT stage of 2 epochs with a learning rate of 1e-5 and batch size of 32. The subsequent DAgger-SFT stage has the same batch size and learning rate, and runs for 1 epoch. Finally, for the RFT stage, we use a batch size of 512 and a learning rate of 5e-6, performing 120 iterations of policy updates.
All experiments are conducted on 4 NVIDIA H800 (80GB) GPUs.
Note that all the experimental results reported below (except for the ablation studies) are obtained by the same fine-tuned model.

\subsection{Main Results}
As shown in Tables~\ref{tab:main-EBALF} and~\ref{tab:main-AW}, our framework of capability-driven planning achieves leading performance on both EB-ALFRED and ALFWorld. Specifically, EB-ALFRED focuses on generalization to diverse instruction styles, including many cases with complex syntax and referential expressions that are rarely seen in the training set.
RoboAgent surpasses all existing fine-tuning-based methods~\cite{RREP,WAP,RoboGPT-R1} in terms of average SR. It also outperforms powerful closed-source models on the base, visual appearance, and long-horizon splits. 
As for ALFWorld, RoboAgent demonstrates an even more significant improvement over existing RL-based methods~\cite{GTR,FTLVLM,GFlowVLM,TCPO,CoSo,SEEA-R1}. The primary reason for this success lies in the improved exploration behavior brought by the high-quality supervision in the SFT stages. In ALFWorld, the agent can only navigate to the receptacles in the room, thus requiring a longer exploration process to locate the target object. Since reward signals are sparse during exploration, it might be difficult for standard RL to learn effective exploration strategies. In contrast, the EG and OG capabilities in our pipeline lead to an explicit exploration process. By providing supervising signals to these modules, the model learns to identify the receptacles that are more likely to contain the target object, resulting in strong SR across all task categories. Visualization of the model's planning process and an analysis of the failure cases can be found in the Appendix.

Apart from the vision-based environment, ALFWorld also provides a text-based simulator, where the observation is a textual description of the objects visible to the agent, and the outcome of the agent's previous action. We adapt the image-related capabilities (OG, SD, ES) to parse textual observations, thereby transforming the multimodal model into a text-only one without additional fine-tuning.
As shown in Table~\ref{tab:main-AW-text}, our VLM agent achieves competitive performance in both seen and unseen environments, reaching success rates comparable to those of recent LLM-based agents that are designed for text-only tasks and built upon larger backbones.
The result indicates that our approach acquires modality-agnostic planning skills, effectively generalizing across visual and linguistic inputs.

Tables~\ref{tab:ood-hab} and~\ref{tab:ood-lota} further present the evaluation results in another visual environment (EB-Habitat) and another textual environment (LoTa-WAH), both built upon simulators different from the one used in training.
These environments differ substantially in terms of object categories, action spaces, and task types.
Compared to other open-source models trained on the ALFRED dataset~\cite{RREP,RoboGPT-R1,LoTa-Bench}, RoboAgent achieves better performance, demonstrating cross-domain generalization to some extent. However, there still remains a noticeable gap between transferred models and the zero-shot, closed-source baselines, suggesting that the domain discrepancy between simulators is still significant.
We hope to address this issue through constructing training data of a larger scale and higher diversity in future work.

To sum up, the proposed capability-driven planning shows a certain degree of generalization across harder \textbf{instructions}, novel \textbf{scenes}, multiple \textbf{modalities}, and different \textbf{task domains}.

\begin{figure}[]
  \centering
  \includegraphics[width=0.86\linewidth]{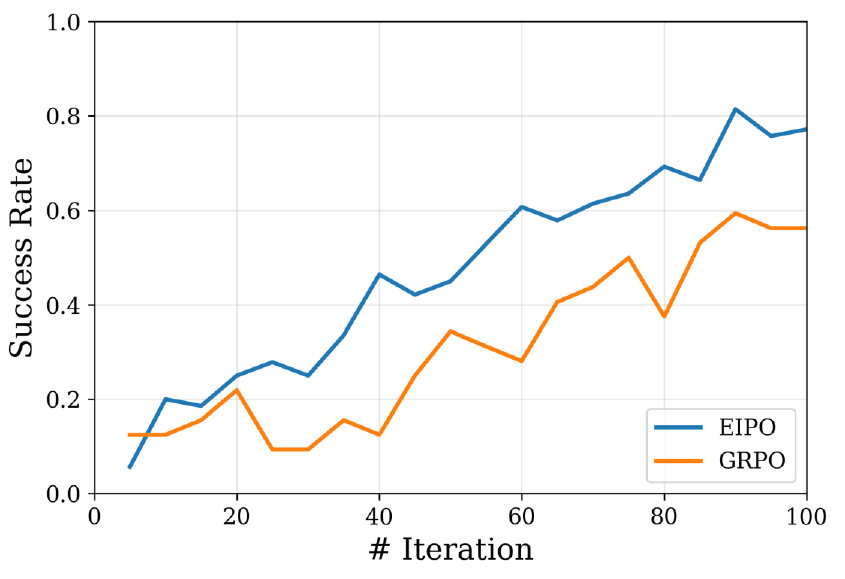}
  \vspace{-3mm}
  \caption{The curve of SR on ALFWorld's val\_seen split during the training process of EIPO and GRPO.}\label{fig:EIPO}
  \vspace{-5mm}
\end{figure}

\vspace{-2mm}
\subsection{Ablation Studies}
\vspace{-1mm}
The ablation results presented in Table~\ref{tab:main-abl} demonstrate the rationality of our proposed training pipeline.
First, SFT on expert trajectories equips the model with fundamental planning-related capabilities and enables it to generate outputs in the desired format, which facilitates the coordination between the scheduler and different capabilities.
Subsequently, performing a second-stage SFT using augmented expert data (aug. exp.) improves the score on EB-ALFRED but leads to a degradation on ALFWorld. This is because EB requires stronger open-ended instruction understanding, while ALFWorld adopts a relatively fixed instruction format and is thus less sensitive to data augmentation.
When further incorporating model-generated data (aug. gen.), the DAgger training yields a significant performance gain on ALFWorld, highlighting the importance of constructing fine-grained corrective supervision for capability learning.
Finally, during the RFT stage, training solely on expert trajectories (exp.) has limited effect. This is because, after the first two stages, the model has already become familiar with the expert dataset, leading to a highly deterministic output distribution. Incorporating data augmentation (aug. exp.) and synthetic capability invocation trajectories (aug. syn.) effectively enhances the scheduler’s performance under more diverse states and complex tasks, resulting in a further improvement over the SFT stages.

In addition, to separately assess the effect of the proposed EIPO algorithm, we conduct an experiment to compare it with GRPO~\cite{Deepseekmath} under controlled conditions. In particular, we fine-tune the same base LLM in the ALFWorld's text environment using either EIPO or GRPO. EIPO leverages step-wise advantages computed under the expert action policy as the optimization target, whereas GRPO uses episode-level returns. As shown in Fig.~\ref{fig:EIPO}, the more stable objective in EIPO helps the model achieve higher SR within the same number of iterations.

\vspace{-5mm}
\section{Conclusion}
\vspace{-3mm}
In this work we propose a capability-driven ETP framework termed RoboAgent, where a VLM serves as both the capability scheduler and five specific capabilities, decomposing the complex planning process into a series of basic vision-language understanding problems. To train the VLM to meet the requirements of each module, we design a three-stage training paradigm that integrates SFT and RFT, as well as expert trajectories, model-generated data, and synthetic data. For supervision and reward design, we make full use of the internal information in the environment simulator, and develop an expert-guided policy optimization algorithm. Experimental results on multiple benchmarks and simulators demonstrate the effectiveness and generality of the proposed approach.
Future directions include scaling up the training data to further improve the generalization of individual capabilities, exploring a dynamic and continually evolving set of capabilities, and extending the proposed framework to a broader class of agentic tasks.

\section*{Acknowledgment}
This work was supported in part by research resources provided through the collaboration between Peking University and XYZ Embodied AI.
{
    \small
    \bibliographystyle{ieeenat_fullname}
    \bibliography{main}
}

\clearpage
\setcounter{page}{1}
\maketitlesupplementary

\section{Details of the Scheduler and Capabilities}
In this section we present a detailed introduction of the input and output format of each VLM calling. Fig.~\ref{fig:prompt-EG} shows the input and output format of the Exploration Guidance (EG) capability. It takes the object query generated by the scheduler as the exploration target, and reads in a candidate list of all the objects in the scene that can serve as the navigation goal (detailed in Sec~\ref{subsec:AW_EB}). In addition, it incorporates a simple memory module that records the historical outputs associated with the same query object, thereby preventing repeated attempts that previously resulted in failure. Whenever EG receives a new query that differs from the previous one, the exploration history is reset to an empty list. The output of EG is a possible location of the target object, which is then passed to AD to be transformed into concrete exploration actions.

Fig.~\ref{fig:prompt-OG} shows the input and output format of the Object Grounding (OG) capability. It follows the standard practice in open-vocabulary grounding tasks and generates JSON-style annotations as output. It also implicitly converts free-form object queries into a clear object category (the ``label" field in the annotations), thereby facilitating subsequent planning of the scheduler. Note that the bounding box coordinates provided by OG are actually not used during inference, as current simulators do not require object location information when performing interactions. We have OG output these coordinates to better supervise its object recognition performance during training, as well as to prepare for future integration with a low-level controller.

Fig.~\ref{fig:prompt-SD} shows the input and output format of the Scene Description (SD) capability. Basically, it translates the image observation into textual representations for the subsequent calling of AD. To ensure that it focuses on the target object to be manipulated, its prompt includes an object query, which is the object category output by the previous OG calling. The output of SD describes the on/in relationships between the target object and other objects, as well as the properties of the target object.

In our implementation of the Action Decoding (AD) capability, we further divide it into two modes: one for exploration sub-plans (Fig.~\ref{fig:prompt-AD-E}) and another for manipulation sub-plans (Fig.~\ref{fig:prompt-AD-M}). The former is responsible for converting the exploration direction output by EG into an action sequence, typically involving only navigation and open actions (when inspecting the inside of a container). The latter converts the manipulation command provided by the scheduler into an action sequence, which requires analyzing the specific state of the target object (for instance, to put \texttt{Apple} to \texttt{Fridge}, it needs to first \texttt{open} the \texttt{Fridge} if \texttt{Fridge} is \texttt{closed}). Therefore, it receives the description generated by SD.
Additionally, during task execution, we record each navigation action and each pick/place action performed by the agent, thereby maintaining the agent’s inventory and location. These two variables are also provided as inputs to the manipulation AD capability to assist in generating atomic actions with the correct object references. 
It should be noted that the prompts of AD are simulator-specific. Fig.~\ref{fig:prompt-AD-E} and~\ref{fig:prompt-AD-M} use ALFWorld~\cite{ALFWorld} as an example. When testing in other simulators, the action space description in the prompts should be adapted accordingly.

Fig.~\ref{fig:prompt-ES} shows the input and output format of the Experience Summarization (ES) capability. The input contains the manipulation command issued by the scheduler, along with the actions and their execution outcomes (success or not) produced by the previous AD calling. The output is a summary of this action history.

Fig.~\ref{fig:prompt-scheduler} shows the input and output format of the scheduler. The prompt contains a description of the 6 predefined capabilities (AD is split into ``exploration\_planner" and ``manipulation\_planner" for easier understanding) and their corresponding arguments. 
The scheduler maintains a memory of historical queries and feedback. We retain only the feedback from OG (whether the target object is found) and ES (whether the manipulation sub-plan is successful). Feedback from other modules (EG and SD) is primarily used by other capabilities within the same scheduler iteration and does not have much impact on subsequent process of the scheduler.
The scheduler’s output consists of two components: the Chain-of-Thought (CoT) and the invoked capabilities. The former includes an analysis of completed and remaining sub-plans based on the history, while the latter comprises a sequence of capability names paired with their corresponding query arguments. We note that the scheduler does not need to provide arguments for AD (exploration) or SD, as their queries come directly from the outputs of the previous EG and OG invocations, respectively.

\begin{figure*}[]
\centering
\resizebox{\textwidth}{!}{
\begin{tcolorbox}[colback=gray!5!white, colframe=black!75!black, 
title=The Input and Output Format of Exploration Guidance (EG), boxrule=0.3mm, width=\textwidth, arc=3mm, auto outer arc=true]
\textbf{Prompt}:

Suppose you are a helpful robotic agent in an indoor environment. Your task is to find \{object query\} in the house, based on common house layouts and object placements. Currently, you can observe the following objects in the house: \{list of available objects\}. You need to output an exploration direction in the exact form of \textless relation\textgreater\ \textless object\textgreater, where \textless relation\textgreater\ is chosen from [target, in, on, near], and \textless object\textgreater\ is an object from the given object list. Previously, you have tried the following exploration directions: \{exploration history\}. Do not output the directions in this list again since they all failed.

\textbf{Output example}:

On CounterTop 1
\end{tcolorbox}
}
\caption{The input and output format of the EG capability.}
\vspace{-1mm}
\label{fig:prompt-EG}
\end{figure*}

\begin{figure*}[]
\centering
\resizebox{\textwidth}{!}{
\begin{tcolorbox}[colback=gray!5!white, colframe=black!75!black, 
title=The Input and Output Format of Object Grounding (OG), boxrule=0.3mm, width=\textwidth, arc=3mm, auto outer arc=true]
\textbf{Prompt}:

\textless image\textgreater

Locate \{object query\} in the image. If you can find it, output the bounding box in json format (including the object class as ``label" of the bounding box); if you cannot find it, output no

\textbf{Output example}:

[\{``bbox\_2d": [193, 104, 308, 253], ``label": ``alarm clock"\}]
\end{tcolorbox}
}
\caption{The input and output format of the OG capability.}
\vspace{-5mm}
\label{fig:prompt-OG}
\end{figure*}

\begin{figure*}[]
\centering
\resizebox{\textwidth}{!}{
\begin{tcolorbox}[colback=gray!5!white, colframe=black!75!black, 
title=The Input and Output Format of Scene Description (SD), boxrule=0.3mm, width=\textwidth, arc=3mm, auto outer arc=true]
\textbf{Prompt}:

\textless image\textgreater

This is an egocentric image observed by a robotic household agent. Please describe the \{object query\} in the scene.

\textbf{Output example}:

There is a laptop on the desk. The laptop is open.
\end{tcolorbox}
}
\caption{The input and output format of the SD capability.}
\vspace{-1mm}
\label{fig:prompt-SD}
\end{figure*}

\begin{figure*}[]
\centering
\resizebox{\textwidth}{!}{
\begin{tcolorbox}[colback=gray!5!white, colframe=black!75!black, 
title=The Input and Output Format of Action Decoding (AD), boxrule=0.3mm, width=\textwidth, arc=3mm, auto outer arc=true]
\textbf{Prompt}:

Suppose you are a helpful robotic agent in an indoor environment. You are able to perform the following actions: 1. go to \textless object\textgreater\  2. open \textless object\textgreater\  3. close \textless object\textgreater\  

The \textless object\textgreater\  is an object name composed of an object class and an object ID (e.g., Apple 1, DiningTable 2).

Now, your task is to \{exploration query\}. Please complete this task by performing one or a series of actions. You should output a list of actions, e.g. [go to Fridge 1], or [go to Cabinet 2, open Cabinet 2]

\textbf{Output example}:

[find a DeskLamp 1]
\end{tcolorbox}
}
\caption{The input and output format of the AD capability for exploration sub-plans (using ALFWorld's action space).}
\vspace{-5mm}
\label{fig:prompt-AD-E}
\end{figure*}

\begin{figure*}[]
\centering
\resizebox{\textwidth}{!}{
\begin{tcolorbox}[colback=gray!5!white, colframe=black!75!black, 
title=The Input and Output Format of Action Decoding (AD), boxrule=0.3mm, width=\textwidth, arc=3mm, auto outer arc=true]
\textbf{Prompt}:

Suppose you are a helpful robotic agent in an indoor environment. You are able to perform the following actions: 1. go to \textless object\textgreater\ 2. open \textless object\textgreater\ 3. close \textless object\textgreater\ 4. use \textless object\textgreater\  (which means to turn on the object) 5. take \textless object\textgreater\  from \textless object\textgreater\  (which means to grasp something from its receptacle) 6. put \textless object\textgreater\  to \textless object\textgreater\  (which means to put something you are holding to a receptacle) 7. cool \textless object\textgreater\  with \textless object\textgreater\  (which means to make something you are holding cold with a tool receptacle) 8. heat \textless object\textgreater\  with \textless object\textgreater\  (which means to make something you are holding hot with a tool receptacle) 9. clean \textless object\textgreater\  with \textless object\textgreater\  (which means to make something you are holding clean with a tool receptacle)

The \textless object\textgreater\  is an object name composed of an object class and an object ID (e.g., Apple 1, DiningTable 2). You need to perform a sequence of actions to complete a given task. You should output a list of actions in the given format, e.g. [take Cup 1 from CounterTop 2], [open Cabinet 1, put Apple 3 to Cabinet 1, close Cabinet 1].

Now, you are holding \{agent inventory\}. You are at \{agent location\}. The environment infomation is: \{scene information\}

Your task is to \{manipulation query\}. Please generate a list of actions in the given format to complete this task.

\textbf{Output example}:

[open the Fridge 1, put down the object in hand, close the Fridge 1, open the Fridge 1, pick up the BreadSliced, close the Fridge 1]
\end{tcolorbox}
}
\caption{The input and output format of the AD capability for manipulation sub-plans (using ALFWorld's action space).}
\vspace{-1mm}
\label{fig:prompt-AD-M}
\end{figure*}

\begin{figure*}[]
\centering
\resizebox{\textwidth}{!}{
\begin{tcolorbox}[colback=gray!5!white, colframe=black!75!black, 
title=The Input and Output Format of Experience Summarization (ES), boxrule=0.3mm, width=\textwidth, arc=3mm, auto outer arc=true]
\textbf{Prompt}:

\textless image\textgreater

Suppose you are a helpful robotic agent in an indoor environment. Your task is to \{manipulation query\}. Here is a list of actions you have performed and their corresponding environment feedbacks: {action history}

Your current egocentric observation is shown in the image. Please summarize the progress made and analyze the reasons for the failures (if any).

\textbf{Output example}:

You successfully grasp SoapBar 1.
\end{tcolorbox}
}
\caption{The input and output format of the ES capability.}
\vspace{-5mm}
\label{fig:prompt-ES}
\end{figure*}

\begin{figure*}[]
\centering
\resizebox{\textwidth}{!}{
\begin{tcolorbox}[colback=gray!5!white, colframe=black!75!black, 
title=The Input and Output Format of the scheduler, boxrule=0.3mm, width=\textwidth, arc=3mm, auto outer arc=true]
\textbf{Prompt}:

Suppose you are a helpful robotic agent in an indoor environment. You have the following abilities and you can invoke them by function calling: 

1. exploration\_guidance(object\_information): given the name or the description of an object, output a direction for exploration 

2. exploration\_planner(): explore the environment accrording to the exploration direction

3. object\_grounding(object\_information): given the name or the description of an object, find it in the egocentric view of the robot

4. scene\_description(): describe the egocentric observation of the robot

5. manipulation\_planner(subtask): given a subtask instruction (a subtask is defined as a part of the task that can be completed within the scene observed from the robot's current egocentric view), complete it by performing atomic actions

6. experience\_summarization(subtask): summarize the previous subtask execution experience

You need to complete a given task by sequentially generating ability queries. When querying the ability of exploration\_guidance, object\_grounding, manipulation\_planner, experience\_summarization, you need to give them a proper input argument. After querying object\_grounding or experience\_summarization, you will get feedbacks of the grounding results or execution results. At each step, you will be given the history of queries you made and feedbacks you received. You need to output your reasoning process after ``Think: ". Then, you need to output your ability query (or queries) after ``Query: ", or simply output ``Stop" if you believe the task is completed.

Now, here comes the task:

Task: \{task instruction\}

History: \{query and feedback history\}

\textbf{Output example}:

Think: The task is ``To chill lettuce and place it on the counter". I am at the beginning of the task. I should: grasp lettuce, chill lettuce, put lettuce to counter. First, I need to find lettuce.

Query: 1. exploration\_guidance(lettuce)

2. exploration\_planner()

3. object\_grounding(lettuce)
\end{tcolorbox}
}
\caption{The input and output format of the scheduler.}
\vspace{-5mm}
\label{fig:prompt-scheduler}
\end{figure*}


\section{Details of the Training Data}
\subsection{The Training Set of ALFRED}
As stated in Sec~\ref{subsec:data}, the training of our model is conducted on the training split of ALFRED~\cite{Alfred}. ALFRED provides 2.4k distinct training tasks, which are grouped into seven categories (pick and place, stack and place, pick two and place, clean and place, heat and place, cool and place, examine in light). Based on the AI2-THOR 2.0~\cite{AI2-THOR} simulator, ALFRED constructs 120 different scenes (30 each for kitchens, bathrooms, bedrooms, and living rooms). Grounding each task to different scenes yields a total of 6.4k task instances. For each task instance, ALFRED generates expert trajectories using the Fast Forward algorithm~\cite{FF} and manually annotates about 3 high-level task descriptions and step-by-step action instructions for each trajectory. The resulting 20k high-level task descriptions serve as the task instructions for the problem of Embodied Task Planning (ETP). Based on these data, we construct our training sets for different training stages, which will be detailed in the following subsections.

\subsection{Stage 1}\label{subsec:data-stage1}
As discussed in Sec.~\ref{subsec:SFT}, the key to constructing the first-stage training data lies in converting expert trajectories into sequences of capability invocations and corresponding queries. To begin with, we pre-process the task instructions provided by ALFRED using a large language model (LLM) (GPT-4.1-mini). Specifically, we feed each instruction along with its corresponding task category into the LLM, prompting it to parse the instruction and extract the key objects involved in the task. For example, for an instruction ``Move the knife from the counter to the microwave table” belonging to the ``pick and place" category, the LLM will output \{object: knife (hint: on the counter), receptacle: microwave table\}.
We note that ALFRED’s human-written instructions contain substantial noise, \textit{i.e.}, the described content does not always match the actual task requirements. Therefore, we manually filter the parsing outputs and discard the instructions whose extracted object descriptions are inconsistent with the real target object categories. This results in a cleaned set of approximately 15k instructions.
In addition, ALFWorld~\cite{ALFWorld} provides some instruction templates for each ALFRED task type (\textit{e.g.}, ``heat some \textless object\textgreater\  and put it in \textless receptacle\textgreater\ " for ``heat and place"). By substituting the placeholders with the target object categories, we generate one additional instruction for each training task instance, expanding the size of the instruction set to 21k. Although these template-based instructions lack the linguistic diversity of human-written ones, they offer precise and unambiguous descriptions of the corresponding task goals.

After processing the instructions, we now turn to the expert trajectories. ALFRED provides a high-level PDDL plan for each training task, which has a similar pattern with the plan in ETP problems. For each human-written instruction, we align its expert plan to the action space of EB-ALFRED, while the plan corresponding to templated instructions are aligned to the action space of ALFWorld (details of the action spaces will be discussed in Sec.~\ref{subsec:AW_EB}). By mixing plans drawn from the two action spaces during training, we hope that the AD capability learns more generalizable action knowledge.

Next, for each training instruction, we decompose its corresponding expert trajectory into alternating exploration and manipulation sub-plans following the procedure described in Sec.~\ref{subsec:SFT}, and further convert the sub-plans into a sequence of capability invocations. The object descriptions extracted by the LLM are then used as queries for EG and OG.
Each manipulation sub-plan is classified to one of several predefined categories (grasp \textless object\textgreater\ , put \textless object\textgreater\  to \textless receptacle\textgreater\ , turn on \textless object\textgreater\ , slice \textless object\textgreater\ , heat \textless object\textgreater\  with \textless tool\textgreater\ , cool \textless object\textgreater\  with \textless tool\textgreater\ , clean \textless object\textgreater\  with \textless tool\textgreater\ , put \textless object\textgreater\  to \textless receptacle\textgreater\  and grasp \textless receptacle\textgreater\ ). By substituting the placeholders with the parsed object descriptions, we obtain the queries for AD and ES. Through this process, the expert plan is transformed into a corresponding sequence of structured capability invocations. Along with each set of invocations, we also construct a corresponding CoT with templates like ``The task is to \{instruction\}. I have already \{done manipulation sub-plans\}. Next, I should \{remaining manipulation sub-plans\}."

To assess the reliability of the transformed capability sequences, we examine whether they can achieve the task goal assuming that all capabilities provide correct outputs. To this end, we construct ``perfect” implementations of each capability using the ground-truth information available in the simulator: perfect SD is built upon the object locations and properties in the scene graph (SG); perfect OG comes from the segmentation masks; perfect AD comes from the decomposed expert plan; perfect ES returns the action feedback given by the simulator. The ground-truth outputs for EG can also be obtained from the SG.
However, we note that accurately predicting the location of an object in a partially observable environment is intrinsically challenging (\textit{e.g.}, finding a knife in a kitchen containing 10+ cabinets and 10+ drawers). To better reflect this difficulty, we inject perturbations into EG's outputs: with a certain probability, it gives a random exploration direction instead of the ground-truth one.
Using these perfect (or perturbed) capability implementations, we attempt to execute the capability invocation sequences in the simulator. Among the 21k instructions, 16k complete successfully. Failures arise from two major sources. First, the ALFWorld/EB-Habitat simulators are not fully aligned with ALFRED, causing the expert plan derived from ALFRED to be occasionally infeasible in these environments. Second, excessive random exploration may sometimes reach the maximum step limit before the plan has been fully executed. We construct the scheduler’s training data using only the task instances whose capability invocation sequence succeeds.

During the verification procedure, we simultaneously record the input and ground-truth output for each capability invocation, generating the training set for each type of capability. In total, we build datasets of size 130k/157k/74k/203k/60k for EG/OG/SD/AD/ES, respectively, with each training sample formatted as shown in Fig.~\ref{fig:prompt-EG}-\ref{fig:prompt-ES}.
For the scheduler, the 16k successful capability invocation sequences provide 179k samples of scheduler callings, each formatted as in Fig.~\ref{fig:prompt-scheduler}. We split all these samples into training and validation sets using an 80/20 ratio, and perform the first expert-SFT stage based on them.

\subsection{Stage 2}
To construct the training data for the second stage, we deploy the model trained in the first stage on the training tasks and record all the capability invocations. Then we follow the pipeline described in Sec.~\ref{subsec:dagger} to generate corrective ground-truth outputs for each invocation. As for data augmentation, we apply the following strategies:
(1) With the help of an LLM (GPT-4.1-mini again), we generate synonyms for every object category appearing in ALFRED, and randomly replace the object IDs in the candidate list of EG's input with these synonyms (\textit{e.g.}, replacing ``Armchair 1" with ``Couch 1").
(2) We generate descriptive phrases for each object category and randomly replace the object queries in OG's input with them (\textit{e.g.}, replacing ``locate the cabinet" with ``locate the tall storage unit with doors").
(3) For each atomic action, we generate several semantically equivalent rephrasings and randomly replace the action names in both the input and output of AD (\textit{e.g.}, replacing ``pick up the \textless object\textgreater" with ``grasp the \textless object\textgreater").
(4) We further use the LLM to produce additional object categories and construct synthetic samples for AD by substituting the original object categories with the newly generated ones (\textit{e.g.}, replacing ``put Apple 1 to Fridge 1” with “put Shirt 1 to WashingMachine 1”).
The goal of these augmentations is to improve the generalization of the capabilities involved. They prevent the model from overfitting to ALFRED’s limited set of object and action names and encourage it to learn the semantic knowledge. In total, we build a dataset of 820k samples for stage 2, and split it with an 80/20 ratio for training and validation.

\subsection{Stage 3}\label{subsec:data-stage3}
The third training stage focuses on improving the performance of the scheduler. Since the scheduler operates on textual inputs/outputs and does not interact with the environment directly, we do not utilize the simulator when constructing the training data for this phase. Instead, we synthesize the feedback for the invoked capabilities.
For each training task, we first obtain its ground-truth capability invocation sequence (Sec.~\ref{subsec:data-stage1}). To construct the scheduler’s input (Fig.~\ref{fig:prompt-scheduler}), we need to generate feedback for every OG and ES calling. With perfect capability execution, each OG feedback would indicate that the target object is found, and each ES feedback would indicate successful manipulation. However, the trained capabilities do not always provide correct outputs.
To model this, we introduce random errors to the capabilities. For each OG invocation, we assign a certain probability of returning a ``target not found" feedback. In such cases, the scheduler need to repeat the previous (EG, AD, OG) invocations until the target is located. Similarly, for each ES invocation, we randomly return a ``failure" feedback and a plausible failure reason with a certain probability. Upon receiving such feedback, the scheduler must roll back to an earlier sub-plan and re-invoke the corresponding capabilities.
Through this process, we augment the original expert capability sequence ($[\{\hat{q}_1^j\}_{j=1}^{\hat{n}_1},...,\{\hat{q}_{T_1}^j\}_{j=1}^{\hat{n}_{T_1}}]$) with an error-recovery mechanism, producing longer and more realistic interaction trajectories ($[\{q_1^j\}_{j=1}^{n_1},f_1,...,\{q_{T_2}^j\}_{j=1}^{n_{T_2}},f_{T_2}]$) that better reflect the actual behaviors of the capabilities. 

We adopt the augmentation strategy proposed in Stage 2 when synthesizing the feedback. In addition, we construct 20 new task categories and about 3k new task instances by combining ALFRED’s 7 original task types (\textit{e.g.}, ``transfer one hot tomato and one cold tomato to the side table"). For each newly created task, we derive its ground-truth capability-invocation sequence by merging the sequences of its constituent tasks, and then synthesize training samples for scheduler using the error-injection procedure described above. To prevent the newly constructed tasks from being too difficult for the scheduler (so that it cannot generate any reasonable outputs during reinforcement learning), we incorporate their expert invocation sequences (without error recovery but including CoT reasoning, approximately 46k samples) into the SFT dataset of Stage 2. These additional scheduler training signals provide a suitable starting point for the Reinforcement Fine-tuning (RFT) phase.

Finally, stage 3 yields 25k task episodes and 360k samples (in the format of Fig.~\ref{fig:prompt-scheduler}, but without CoT), which are again split into training and validation sets using an 80/20 ratio. The details of the RFT training based on these samples are presented in Sec.~\ref{sec:RL}.

\section{Details of the Benchmarks}
\subsection{ALFWorld and EB-ALFRED}\label{subsec:AW_EB}
ALFWorld~\cite{ALFWorld} and EB-ALFRED~\cite{EMBODIEDBENCH} are used as the primary evaluation environments.
The ALFWorld benchmark provides a wrapper of ALFRED’s low-level actions and implements a text-based action interface through the TextWorld~\cite{TextWorld} engine. It follows ALFRED’s original data split. The training set retains 3.6k out of ALFRED’s 6.4k training tasks (removing all tasks in the ``stack and place category" as well as those involving the ``slice" action). Its evaluation set is drawn from ALFRED’s validation split and contains 134 tasks in unseen scenes and 140 tasks in seen scenes during training (both with novel object initialization).
ALFWorld supports two agent modes. The first is a vision-based setting, in which the agent receives an egocentric image (with a default resolution of $300\times300$) at each time step, mirroring the original ALFRED environment.\footnote{Some previous works on ALFWorld's vision-based environment assumes a textual feedback of each action (which may contain a list of newly observed objects), while we only leverage a binary signal (whether the action is available or not) as action feedback during inference.} The second is a text mode, in which the agent receives a textual description of the outcome of the previous action along with the set of currently observable objects. The maximum number of action steps per episode is set to 50. 

EmbodiedBench~\cite{EMBODIEDBENCH} integrates 4 simulation environments to provide a comprehensive evaluation of embodied agents across planning, navigation, and manipulation competencies. Among them, EB-ALFRED is derived from the LoTa-ALFRED~\cite{LoTa-Bench} benchmark and introduces a wrapper for the ALFRED environment that differs from that of ALFWorld. It offers 300 evaluation tasks, all sourced from ALFRED’s validation set of seen scenes. These tasks span all 7 task categories of ALFRED, and are reorganized into 6 splits (base, visual appearance, spatial relationship, complex instruction, common sense, long horizon) to capture distinct characteristics of the task instructions.
EB-ALFRED adopts a setting of vision-based observation with a default resolution of $500\times500$. The maximum number of action steps per episode is set to 30, and an episode also terminates if the agent outputs 10 invalid actions.

Table~\ref{tab:aw_eb} summarizes several key differences between ALFWorld and EB-ALFRED. Specifically:
(1) \textbf{Task diversity}. EB-ALFRED includes a richer set of task types than ALFWorld, \textit{e.g.}, ``put \textit{a bowl with a knife in it} to the countertop", ``put a \textit{sliced} apple to the countertop".
(2) \textbf{Action granularity}. In ALFWorld, ``heat", ``cool", and ``clean" are implemented as atomic actions. In contrast, EB-ALFRED requires the agent to realize these effects through more primitive operations (\textit{e.g.}, \textit{cooling} an apple by \textit{putting} it to the fridge and then \textit{picking} it up). EB-ALFRED also provides a ``slice" action, which ALFWorld lacks because it contains no tasks involving sliced objects.
(3) \textbf{Navigation behavior}. ALFWorld restricts navigation to ``go to” actions targeting large, immovable receptacles. EB-ALFRED, however, allows the agent to approach any object in the environment via a ``find" action. This makes ALFWorld place greater emphasis on exploration: the agent must reason about the potential locations of the target object in order to find it.
(4) \textbf{Instruction design}. ALFWorld uses templated instructions in which the target object category is explicitly mentioned. In contrast, EB-ALFRED uses human-written (or GPT-augmented) instructions that are longer, syntactically richer, and often refer to target objects more indirectly (\textit{e.g.}, by describing their appearance or location instead of naming their category).
(5) \textbf{Evaluation environments}. Assuming that the training is performed on ALFRED’s training tasks, then ALFWorld evaluates the agent in both seen and unseen scenes during training, whereas EB-ALFRED's evaluation is conducted in seen scenes.
Overall, EB-ALFRED emphasizes task diversity and instruction complexity, while ALFWorld places more focus on efficient exploration and generalization to novel scenes.

Our goal is to train a model that can bridge the aforementioned discrepancies and operate seamlessly across ALFWorld and EB-ALFRED. To achieve this, as described in Sec.~\ref{subsec:data-stage1}, we use ALFRED's training set (which is a superset of ALFWorld's training set) and include both the human-annotated instructions from ALFRED and the templated instructions from ALFWorld. To address the differences in atomic actions, we provide AD with training data that encompass different action spaces. To handle the differences in navigation behavior, we construct two variants of the candidate list in the EG input: a “receptacle-only” setting and an “all-objects” setting. Both the receptacle list and the object list can be extracted from the set of available actions.

\begin{table*}
  \caption{A comparison between ALFWorld~\cite{ALFWorld} and EB-ALFRED~\cite{EMBODIEDBENCH}. We slightly modify the names of the atomic actions (\textit{e.g.}, replacing ``find" with ``go to") to enable a clearer comparison of the differences.}
  \label{tab:aw_eb}
  \centering
  \scalebox{0.98}{
  \begin{tabular}{cccccc}
    \toprule
      & Task Types & Action Space & Nav. Target & Instruction & Test Scene \\
      \midrule
ALFWorld & 6 & go to, pick, put, turn on, open, close, & only receptacles & template-based & seen\&unseen\\
& & heat, cool, clean\\
    \midrule
  EB-ALFRED & 7 & go to, pick, put, turn on, open, close,  & all objects & human-written & seen \\
  & & slice \\
  \bottomrule
  \end{tabular}
  }
  \vspace{-1mm}
\end{table*}

\subsection{Adaptation to EB-Habitat}
EB-Habitat is another benchmark for ETP designed by EmbodiedBench~\cite{EMBODIEDBENCH}. It is constructed upon the Language Rearrangement task proposed by LLaRP~\cite{LLMGP}. EB-Habitat adopts the same observation format as EB-ALFRED, and its 300 testing tasks are partitioned into the same 6 splits to capture the diversity of instruction styles. However, since EB-Habitat and EB-ALFRED are built on different simulators with substantially different sets of object categories and task types, it serves as a challenging out-of-distribution (OOD) evaluation environment for our trained model.

When evaluating on EB-Habitat (Table~\ref{tab:ood-hab} of the main paper), we keep the VLM parameters fixed and introduce the following modifications for the capability invocation pipeline. For AD, the action descriptions in the prompt (Figs.~\ref{fig:prompt-AD-E} and~\ref{fig:prompt-AD-M}) are replaced with the action space of EB-Habitat (``navigate", ``pick", ``place", ``open", ``close"). Similar to ALFWorld, EB-Habitat only permits navigation actions toward receptacles, so we set the candidates in the prompt of EG to the list of receptacles present in the scene.
In practice, we find that the primary obstacles to OOD generalization stem from the visual domain: EB-Habitat provides less realistic environment rendering than ALFRED and also contains many object categories that never appear in ALFRED. As a result, capabilities that rely on image inputs exhibit degraded performance. To mitigate this issue, we introduce several additional adjustments. Noting that EB-Habitat involves relatively few object-state changes and that action failures occur in a highly similar pattern (mostly ``target object out of reach”), we omit SD (\textit{i.e.}, make it return an empty string when invoked) and remove the image input from ES.
For OG, the agent first checks whether the object query belongs to the set of object categories provided by EB-Habitat (which can be parsed from the set of valid actions). If it does, the model directly returns that the target object is found. If it does not, the model performs the standard open-vocabulary grounding procedure, after which the detected object label is mapped to the most similar string in the category set. This adjustment compensates for the model’s limited recognition ability at the cost of increased number of action steps (\textit{e.g.}, attempting to grasp an apple even when it is not visible now).\footnote{We also observe that EB-Habitat occasionally presents situations where the agent is close to an object but the object is not visible in the rendered view, which further motivates the above modification to the OG capability.}

\subsection{Adaptation to Text Environments}
In Tables~\ref{tab:main-AW-text} and~\ref{tab:ood-lota} of the main paper, we also evaluate the model on two text environments: ALFWorld-Text and LoTa-WAH. ALFWorld-Text is already introduced in Sec.~\ref{subsec:AW_EB}. To adapt the VLM agent to receive textual observations, we make the following adaptations to the capabilities. (1) The candidate list of EG's input is parsed from the initial observation, which lists all the receptacles in the room. (2) OG is implemented by checking whether the queried object appears in the object list extracted from the most recent observation. (3) SD is implemented by outputting ``There is a \{object query\}” and appending the property descriptions extracted from the latest observation (if any). (4) The image input for ES is removed. Comparing the results in Tables~\ref{tab:main-AW} and~\ref{tab:main-AW-text}, we observe that the ETP task is easier in the text-based environment than in the visual environment. This indicates that visual recognition of objects and their properties remains one of the key factors limiting planning performance. This is also validated by the error analysis in Sec.~\ref{sec:EA}.

LoTa-Bench~\cite{LoTa-Bench} is another text-based benchmark for embodied planning. It organizes two simulators for evaluation, ALFRED~\cite{Alfred} and VirtualHome~\cite{VirtualHome}. Here, we focus on the latter, LoTa-WAH, since EB-ALFRED basically subsumes the former. Compared with ALFWorld-Text, LoTa-WAH exhibits a different set of object categories and does not provide updated environment observations. Instead, it only reports the complete list of object categories at the beginning of a task and indicates whether each executed action succeeds or fails at each timestep.
Given these characteristics, we omit SD (which directly returns an empty string) and OG (which directly returns ``the target object is found”), and remove the image input from ES. For EG, we note that LoTa-WAH allows navigation actions toward any object, but the number of objects in the scene is large. Accordingly, we retain only all receptacles and the objects having high semantic similarity with the object query (measured using all-MiniLM-L6-v2~\cite{Sentence-BERT}) in the input candidate list. For AD, we modify the prompt to match the action space defined in LoTa-WAH. LoTa-WAH uses subgoal success rate (SSR) as the evaluation metric, defined as the proportion of the predefined goal conditions that are satisfied at the end of an episode.

\section{Preliminaries on the RFT Stage}\label{sec:RL}
Reinforcement learning (RL) aims to learn a policy $\pi$ that generates an action distribution given a state: $a_t\sim\pi(\cdot|s_t)$. After taking each action, it will receive a reward $R(s_t,a_t)$. The most classical training objective is the expected return of the policy:
\begin{equation}
    \eta(\pi)=\mathbb{E}_{(s_0,a_0,...,s_T,a_T)\sim\pi}[\sum_{t=0}^T\gamma^tR(s_t,a_t)],
\end{equation}
where $\gamma$ is a decay factor, $T$ is the maximum length of the episode, $\tau=(s_0,a_0,...,s_T,a_T)$ is a trajectory collected by rolling out $\pi$ in the environment. With a parameterized policy $\pi_{\theta}$, Policy Gradient~\cite{PG} shows that the gradient of $\eta$ can be computed with:
\begin{equation}
    \nabla_{\theta}\eta(\pi)=\mathbb{E}_{(s_0,a_0,...,s_T,a_T)\sim\pi}[\sum_{t=0}^T\nabla_{\theta}\log\pi_{\theta}(s_t|a_t)A_{\pi}(s_t,a_t)],
\end{equation}
where the advantage $A_{\pi}(s_t,a_t)=Q_{\pi}(s_t,a_t)-V_{\pi}(s_t)$, and $Q_{\pi}(s_t,a_t)=R(s_t,a_t)+\gamma V_{\pi}(s_{t+1})$, $V_{\pi}(s_t)=\mathbb{E}_{(a_t,s_{t+1}...,s_T,a_T)\sim\pi}[\sum_{t'=t}^T\gamma^{t'-t}R(s_{t'},a_{t'})]$.
However, this limits the policy training to using strict on-policy data ($(s_0,a_0,...,s_T,a_T)\sim\pi$). To allow gradient optimization with slightly off-policy data, \cite{AOARL} shows that:
\begin{equation}
    \eta(\pi')=\eta(\pi)+\mathbb{E}_{(s_0,a_0,...,s_T,a_T)\sim\pi'}\sum_{t=0}^T\gamma^tA_{\pi}(s_t,a_t).\label{eq:AOARL}
\end{equation}
TRPO further gives a surrogate for the right hand side:
\begin{align}
\eta(\pi')-\eta(\pi) &=\mathbb{E}_{s\sim d_{\pi'}}\mathbb{E}_{a\sim\pi'(\cdot|s)}A_{\pi}(s,a) \\
&\approx\mathbb{E}_{s\sim d_{\pi}}\mathbb{E}_{a\sim\pi(\cdot|s)}\frac{\pi'(a|s)}{\pi(a|s)}A_{\pi}(s,a).
\label{eq:TRPO}
\end{align}
where $d_{\pi}$ is the (unnormalized) state distribution under $\pi$: $d_{\pi}(s)=P(s_0 = s)+\gamma P(s_1 = s)+\gamma
^2P(s_2 = s)+...+\gamma^TP(s_T=s)$. 
Now, $\pi'$ in Eq (\ref{eq:TRPO}) can be optimized using trajectories generated with an old policy $\pi$. TRPO implements this by performing constraint optimization to avoid $\pi'$ from deviating too far from $\pi$, while PPO~\cite{PPO} achieves this through a clipping mechanism:
\begin{align}
    J_{\text{PPO}}(\pi')=\mathbb{E}_{s\sim d_{\pi},a\sim\pi(\cdot|s)}\min[r(a,s)A_{\pi}(s,a), \notag \\
    \text{clip}(r(a,s),1+\epsilon,1-\epsilon)A_{\pi}(s,a)],
\end{align}
where $r(a,s)=\frac{\pi'(a|s)}{\pi(a|s)}$ is adaptively clipped according to the signal of $A_{\pi}$. PPO calculates the advantage value by General Advantage Estimation~\cite{GAE}, which involves the training of a value model along with the policy. GRPO~\cite{Deepseekmath} further eliminates the need for this value model by using the group-wise average return as the baseline:
\begin{align}
    J_{\text{GRPO}}(\pi')=\mathbb{E}_{s\sim d_{\pi},a^i\sim\pi(\cdot|s)}\sum_{i=1}^G\min[r(a,s)\hat{A}_{\pi}(s,a^i), \notag \\
    \text{clip}(r(a,s),1+\epsilon,1-\epsilon)\hat{A}_{\pi}(s,a^i)],
\end{align}
where $\hat{A}_{\pi}(s,a^i)$ is the return of action $a^i$ normalized with the group's mean and standard deviation.

Our proposed Expert-Induced Policy Optimization (EIPO) follows Eq (\ref{eq:AOARL}) but substitute $\pi$ with an expert policy $\pi^*$. So, the optimization target is transformed to:
\begin{align}
    \eta(\pi')-\eta(\pi^*)&=\mathbb{E}_{s\sim d_{\pi'}}\mathbb{E}_{a\sim\pi'(\cdot|s)}A_{\pi^*}(s,a) \label{eq:EIPO_ori} \\ 
&\approx\mathbb{E}_{s\sim D}\mathbb{E}_{a\sim\pi(\cdot|s)}\frac{\pi'(a|s)}{\pi(a|s)}A_{\pi^*}(s,a),\label{eq:EIPO_D}
\end{align}
where $D$ is a static dataset of policy input. By introducing $\pi^*$, EIPO bypasses the estimation of returns and state values under policy $\pi$ through value models or Monte Carlo sampling. Instead, it adopts the more stable expert advantage function $A_{\pi^*}$ as the optimization objective. This approach more clearly reflects the credit of each action and provides more effective gradient signals consequently. Furthermore, as discussed in Sec.~\ref{subsec:RFT}, we incorporate PPO-style probability ratio clipping and GRPO-style group normalization into EIPO. Note that GRPO's normalization uses the group mean as an estimate of the state value under policy $\pi$ to approximate the advantage $A_{\pi}(s,a^i)$. In contrast, EIPO's normalization employs the group mean as a baseline for $A_{\pi^*}$ to reduce the variance of the policy gradient and introduce positive gradient signals (detailed in Sec.~\ref{subsec:RFT}). Following recent works like~\cite{DrGRPO}, we omit the standard deviation normalization used in GRPO and only subtract the group average.

The proposed EIPO can be applied to any policy as long as an expert $\pi^*$ can be formulated. We present two examples in this work. The first is a conventional action policy, in which the model directly outputs atomic actions. The second is the scheduler policy in our capability-driven planning framework, where the model outputs capability invocations. For the action policy, we train the model using online data. In each iteration, the model rolls out $G=8$ complete trajectories in each of the batch\_size=16 randomly selected training environments. During planning, we maintain the lists of completed and remaining sub-plans for each timestep. For each state $s$ along each trajectory, suppose the number of remaining sub-plans is $n_s$, then an expert action policy should be able to complete the task within $n_s$ rounds of output (where each output is a list of actions completing a sub-plan). Assuming that completing each sub-plan yields a reward of $+1$, the corresponding expert value for that state, $V_{\pi^*}(s)$, can thus be computed as $\sum_{i=0}^{n_s-1}\gamma^i$. Meanwhile, by examining the difference in the number of remaining sub-plans between consecutive states along the trajectory, we can determine the immediate reward $R(s,a)$ for each action $a$. Based on the reward for $a$ and the expert value for $s$, we obtain the advantage value for each state-action pair by $A_{\pi^*}(s,a)=\gamma V_{\pi^*}(s')+R(s,a)-V_{\pi^*}(s)$, where $s'$ is the next state in the trajectory. After group-based normalization, this is used as the objective for policy gradient and model update. The training is conducted for 150 iterations with an initial learning rate of 1e-6. We implement the online training pipeline on ALFWorld's text environment using the framework provided by GiGPO~\cite{GiGPO}. The experimental results are presented in Sec.~\ref{subsec:abl_RFT}.

In our training stage 3, we apply the EIPO algorithm for the scheduler. Treating the scheduler VLM as a policy, we define the state $s_t$ as all previously generated queries and the corresponding feedback up to step $t-1$, and the action $a_t$ as the sequence of queries produced at the current scheduler calling. To avoid the substantial time cost of policy rollouts during training (which involves image rendering and capability invocations), we sample input states from an offline dataset, as shown in Eq (\ref{eq:EIPO_D}). This deviates from the original objective in Eq (\ref{eq:EIPO_ori}), which ideally samples states from the distribution induced by the current policy $\pi$. However, the trajectory synthesis process described in Sec.~\ref{subsec:data-stage3} enables us to generate a large and diverse dataset (covering scenarios where capability invocations succeed or fail), which helps mitigate the problem of distribution shift to some extent.
The training proceeds for 120 iterations. In each iteration, we sample a batch of 512 states to serve as prompts. For each prompt, the model generates $G=8$ responses, each containing both the CoT reasoning and the capability queries (as in Fig.~\ref{fig:prompt-scheduler}). 
The computation of $A_{\pi^*}$ follows a procedure similar to that used for the action policy. We again maintain the lists of completed and remaining sub-plans for each state, and assume that an expert scheduler would be able to resolve the remaining $n_s$ sub-plans with $n_s$ rounds of output. The reward scheme assigns $+1$ for the completion of each manipulation sub-plan. This choice is motivated by the observation that the successful execution of exploration sub-plans often depends heavily on the performance EG and OG capabilities; thus even a correct scheduler output may not directly complete an exploration sub-plan in practice. Given this reward definition, we can compute $V_{\pi^*}$ for any input state $s$. However, in the offline setting, obtaining the next state $s'$ resulting from an action is challenging, since no environment or capability interaction is available. To estimate $R(s,a)$ and $V_{\pi^*}(s')$, we adopt a simple approximation: when the scheduler’s output matches that of the expert scheduler, we follow the expert policy to determine $s'$ (a state where a new sub-plan is completed). When the output does not match, we assume that the number of remaining sub-plans stays unchanged and the immediate reward is 0. We find that this approximation yields satisfactory empirical performance (as in Table~\ref{tab:main-abl}), and leave the incorporation of a world model and the design of a more delicate reward mechanism for future work. 
After obtaining $A_{\pi^*}$ for each state-action pair (\textit{i.e.}, prompt-output pair), we perform the computation of policy gradient and the update of model parameters following the same procedure as standard GRPO.

\section{More Experimental Results}
\subsection{Impact of Training Data}\label{subsec:abl_data}

Our proposed training pipeline leverages all tasks in ALFRED's training set to develop an agent capable of operating in both the ALFWorld and EB-ALFRED environments. Since ALFWorld uses only a subset of ALFRED's tasks, we re-run the 3-stage training procedure using only ALFWorld's training tasks, in order to enable a more controlled comparison with previous baselines on ALFWorld. The results are presented in Table~\ref{tab:only_aw}. Without additional tasks, Our model still reaches a success rate of 67.2\% after the 3-stage optimization, which is better than all previous methods in Table~\ref{tab:main-AW}. On the other hand, by comparing with the results in Table~\ref{tab:main-abl}, we observe that incorporating the additional ALFRED tasks, despite that they belong to task categories different from ALFWorld, yields a roughly 10\% performance improvement. This suggests that our capability-based framework is able to acquire generalizable planning strategies from a diverse set of tasks and may possess promising scalability. Moreover, introducing synthetic tasks (constructed by combining the original ALFWorld tasks) for EIPO also leads to performance gains, further demonstrating the effectiveness of our data collection strategy in Stage 3.

\subsection{Impact of the Capability-Driven Framework}\label{subsec:abl_cap}

To further demonstrate the advantage of the capability-driven planning framework, we compare it with a plain planner that directly generates actions. Specifically, this baseline uses the same Qwen2.5-VL-3B backbone as our method. At each timestep, it receives the current observation image, the task instruction, and the interaction history (including past actions and the environment feedback of success/failure), and outputs a CoT reasoning process followed by one or more actions. This can be viewed as an integration of the scheduler and AD capability in our framework.
To train this model, we adopt a two-stage pipeline. First, SFT is performed using expert trajectories. The trained model is then utilized to collect trajectories on the training tasks, which serves as the dataset for an EIPO-based RFT stage. 
Here, we adopt an approach analogous to the scheduler policy described at the end of Sec.~\ref{sec:RL} to compute the number of remaining sub-plans and the $V_{\pi^*}$ value for each state along the trajectory. We then construct a set of optimal actions for each state based on the expert scheduler and expert AD. During training, we determine the next state $s'$ by checking whether the model’s predicted action falls within the optimal action set, which then allows us to compute the advantage $A_{\pi^*}$ for optimization.
Note that we omit the DAgger-SFT stage, as it is designed for enhancing the capabilities. Similar to Sec.~\ref{subsec:abl_data}, the training is conducted on ALFWorld's training tasks, and the evaluation results are reported in Table~\ref{tab:plan}.
Under both the SFT-only and SFT+RFT settings, the planner equipped with capabilities consistently outperforms the version that directly generates actions. We observe that the planner without capabilities tends to more frequently produce invalid actions, misinterpret the task progress (\textit{e.g.}, attempting to heat an object despite having failed to pick it), and fail to locate the target object due to insufficient exploration or inaccurate recognition. In addition, the results in Table~\ref{tab:plan} also demonstrate the generality of the proposed EIPO algorithm across different types of planners, which will be elaborated more concretely in the next subsection.

The benefits of introducing capabilities can be understood from two aspects. First, it enables a clearer and more reliable reasoning process, such as the explicit exploration and object localization. Second, it allows the incorporation of additional fine-grained supervisory signals into the training process.

\begin{table}
  \caption{The results of solely using ALFWorld for training.}
  \vspace{-2mm}
  \label{tab:only_aw}
  \centering
  \scalebox{0.96}{
  \begin{tabular}{cccc}
    \toprule
     SFT-expert & SFT-DAgger & EIPO & SR \\
      \midrule
\ding{51} & \ding{55} & \ding{55} & 42.5 \\
\ding{51} & \ding{51} & \ding{55} & 59.7 \\
\ding{51} & \ding{51} & w.o. syn. task & 67.2 \\
\ding{51} & \ding{51} & w. syn. task & 68.6\\
  \bottomrule
  \end{tabular}
  }
  \vspace{-2mm}
\end{table}

\begin{table}
  \caption{The results of an VLM planer with/without explicit capability invocation.}
  \vspace{-2mm}
  \label{tab:plan}
  \centering
  \scalebox{0.96}{
  \begin{tabular}{cccc}
    \toprule
     Cap. & SFT & EIPO & SR \\
      \midrule
\ding{55} & \ding{51} & \ding{55} & 40.3 \\
\ding{55} & \ding{51} & \ding{51} & 48.5 \\
\midrule
\ding{51} & \ding{51} & \ding{55} & 59.7 \\
\ding{51} & \ding{51} & \ding{51} & 68.6 \\
  \bottomrule
  \end{tabular}
  }
  \vspace{-5mm}
\end{table}

\subsection{Impact of the RFT Algorithm}\label{subsec:abl_RFT}
To more thoroughly analyze the effect of the EIPO algorithm, we compare it with the GRPO~\cite{Deepseekmath} baseline under two settings: an online action policy and an offline scheduler policy. For the action policy, the results in Table~\ref{tab:EIPO_GRPO_T} extend those shown in Fig.~\ref{fig:EIPO} of the main text. As described in Sec.~\ref{sec:RL}, we train models within the verl-agent framework provided by GiGPO~\cite{GiGPO}, using Qwen2.5-1.5B-Instruct as the base model and keeping all hyperparameters consistent. Experiments are conducted in the text-based ALFWorld environment.
For a complete episode collected by the model, GRPO assigns a reward solely based on the final signal of task success or failure. This terminal reward is shared by all steps within the episode and used as the optimization target. GiGPO further computes a per-step return for each action and constructs both step-level and episode-level groups for normalization.
In contrast, our method directly uses the expert advantage $A_{\pi^*}$ for each action as the optimization target. We also experiment with constructing both episode-level and step-level normalization groups, analogous to GiGPO. The results demonstrate that the more stable and accurate optimization target of expert advantage enables EIPO to outperform GRPO, which relies on estimating policy returns through sampled rollouts. Moreover, incorporating step-level grouping does not yield additional performance gains. A possible explanation is that EIPO’s per-action advantage computation already addresses much of the credit assignment problem, reducing the need for fine-grained normalization through additional grouping.

For the scheduler, GRPO based on episode-level returns is difficult to apply because we employ an offline dataset to avoid expensive rollouts. As a baseline for EIPO, we consider an alternative algorithm that uses per-action rewards as the optimization target: the model receives a reward of 1 if its output matches the expert scheduler’s output, and 0 otherwise. This approach resembles GRPO on single-turn QA datasets, ignoring the long-term consequences of actions and the multi-turn planning process.
As shown in Table~\ref{tab:EIPO_GRPO}, EIPO achieves consistently better performance than this reward-based baseline in both evaluation environments. We also note that the gain is particularly pronounced on longer-horizon tasks (\textit{e.g.}, the long horizon split of EB-ALFRED and the exploration-intensive tasks in ALFWorld). This indicates that EIPO provides a practical and effective objective for offline policy optimization of VLM-based agents.

\begin{table}
  \caption{A comparison between GRPO and EIPO for training LLM-based (Qwen2.5-1.5B) planner on ALFWorld. The reported results are averaged across 3 independent runs.}
  \vspace{-2mm}
  \label{tab:EIPO_GRPO_T}
  \centering
  \scalebox{0.96}{
  \begin{tabular}{cc}
    \toprule
     objective & SR \\
      \midrule
rollout return (episode-level group) & 72.8±3.6\\ 
rollout return (episode\&step-level group) & 86.7±1.7 \\
expert advantage (episode-level group) & \textbf{94.8}±3.9 \\
expert advantage (episode\&step-level group) & 92.7±3.9 \\
  \bottomrule
  \end{tabular}
  }
  \vspace{-1mm}
\end{table}

\begin{table}
  \caption{A comparison between GRPO and EIPO for the training stage 3 of our pipeline.}
  \vspace{-2mm}
  \label{tab:EIPO_GRPO}
  \centering
  \scalebox{0.96}{
  \begin{tabular}{ccc}
    \toprule
     objective & ALFWorld & EB-ALFRED \\
      \midrule
none & 73.1 & 64.3 \\ 
reward & 69.4 & 65.0 \\
expert advantage & \textbf{77.6} & \textbf{67.0} \\
  \bottomrule
  \end{tabular}
  }
  \vspace{-5mm}
\end{table}

\subsection{Efficiency}
Introducing capability invocations into the planning process may raise concerns regarding computational efficiency. To assess this, we execute our capability-driven planning pipeline for 3 times on 6 ALFWorld tasks (one from each task category) and report the average statistics in Table~\ref{tab:eff}. For comparison, we also evaluate the planner that directly generates actions (described in Sec.~\ref{subsec:abl_cap}) under the same experimental conditions.
All results are obtained on a single NVIDIA RTX 4090.

The results show that incorporating capabilities indeed increases the number of VLM calls. However, the per-call token cost is lower, as many capabilities have short prompts and concise outputs. The overall runtime of the two methods is comparable. It is worth noting that the runtime depends not only on the number of VLM calls but also on the number of interactions with the simulator. For many complex tasks, action-based approaches may spend a large number of tokens and considerable time on blind exploration or invalid actions, which usually result in task failure. In such cases, the capability-based approach offers advantages in both performance and efficiency.

\begin{table}
  \caption{Statics of VLM inference for capability-based and action-based planner.}
  \vspace{-2mm}
  \label{tab:eff}
  \centering
  \scalebox{0.92}{
  \begin{tabular}{ccccc}
    \toprule
      & time & \# call & \# token in & \# token out \\
      \midrule
action-based & 35.6s & 16.3 & 11.3k & 1.1k \\ 
capability-based & 36.3s & 35.1 & 10.1k & 1.1k\\
  \bottomrule
  \end{tabular}
  }
  \vspace{-5mm}
\end{table}

\subsection{Plan visualizations}
\begin{figure*}[h]
  \centering
  \includegraphics[width=0.99\linewidth]{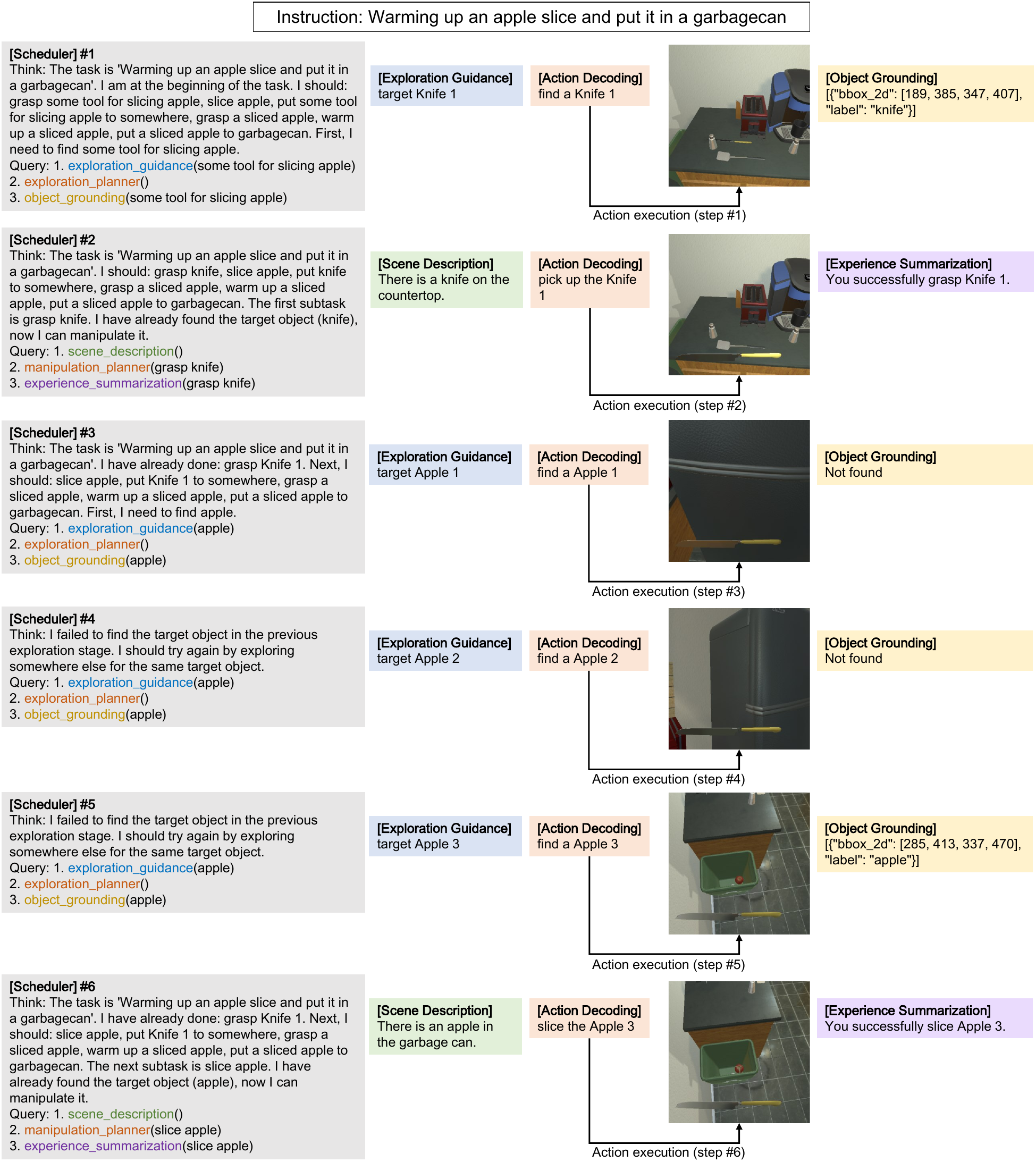}
  \caption{A visualization of the generated plan on EB-ALFRED (long horizon split), part 1.}\label{fig:vis_eb1-1}
  \vspace{-5mm}
\end{figure*}

\begin{figure*}[h]
  \centering
  \includegraphics[width=0.99\linewidth]{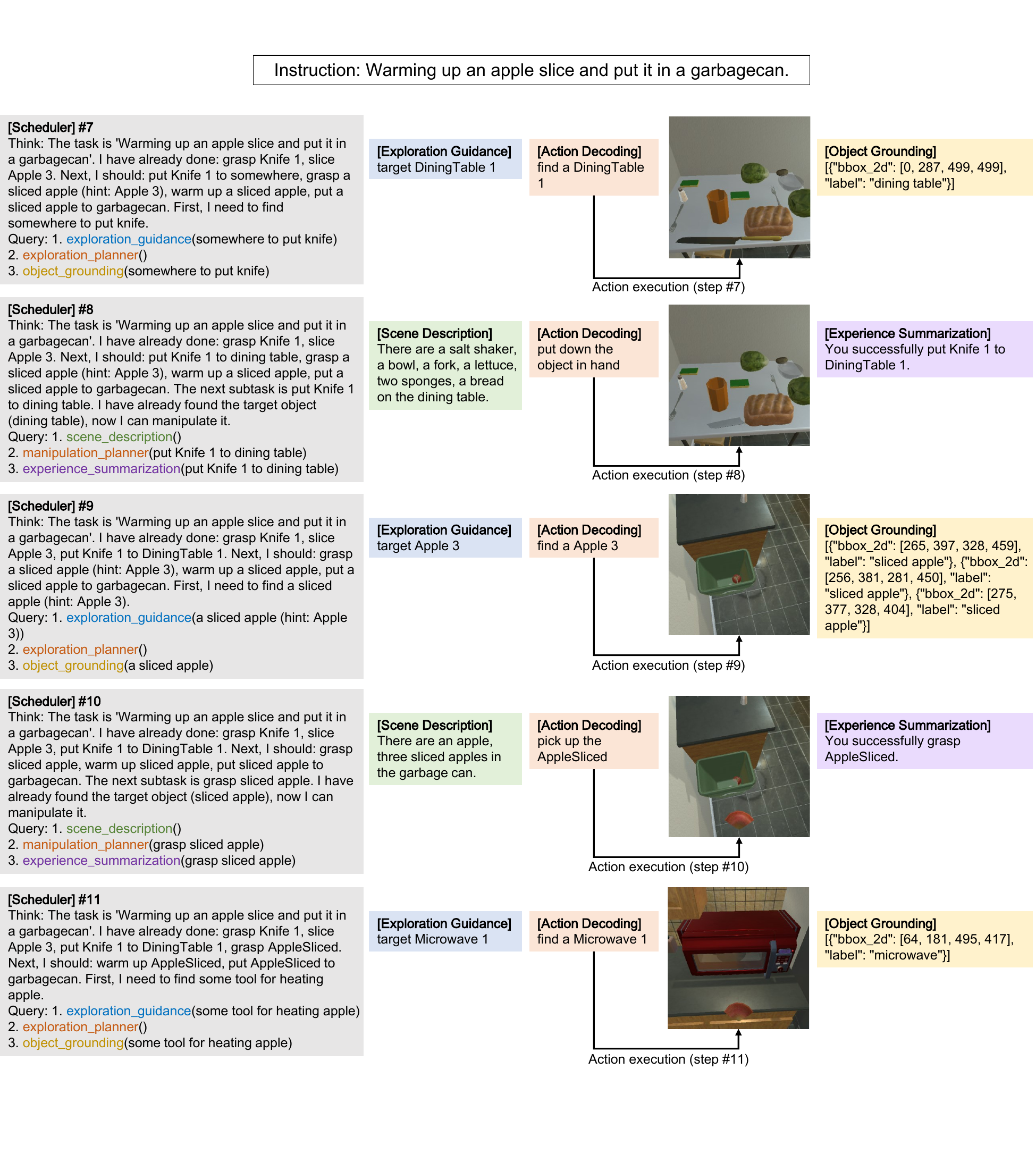}
  \caption{A visualization of the generated plan on EB-ALFRED (long horizon split), part 2.}\label{fig:vis_eb1-2}
  \vspace{-5mm}
\end{figure*}

\begin{figure*}[h]
  \centering
  \includegraphics[width=0.99\linewidth]{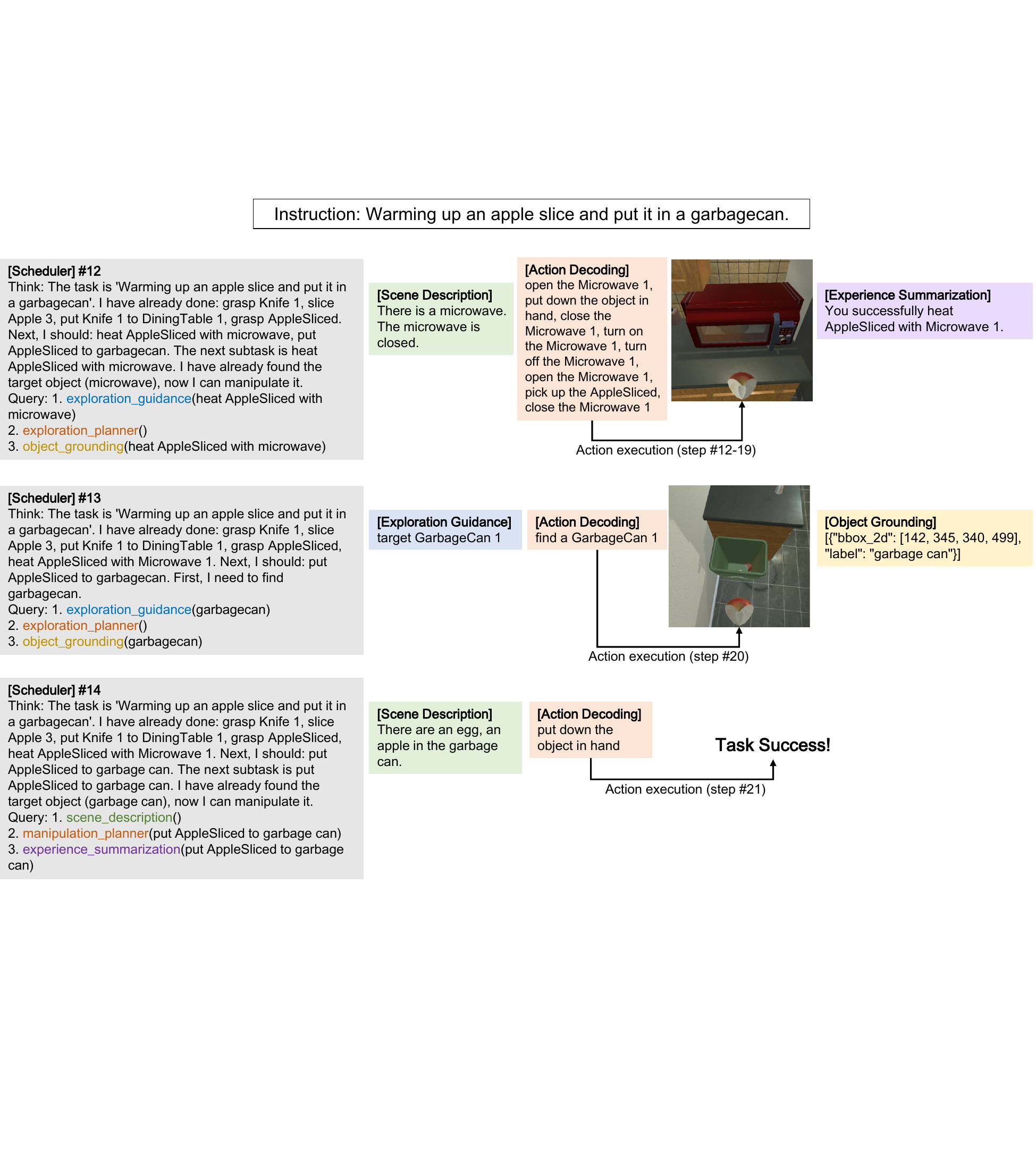}
  \caption{A visualization of the generated plan on EB-ALFRED (long horizon split), part 3.}\label{fig:vis_eb1-3}
  \vspace{-5mm}
\end{figure*}

\begin{figure*}[h]
  \centering
  \includegraphics[width=0.99\linewidth]{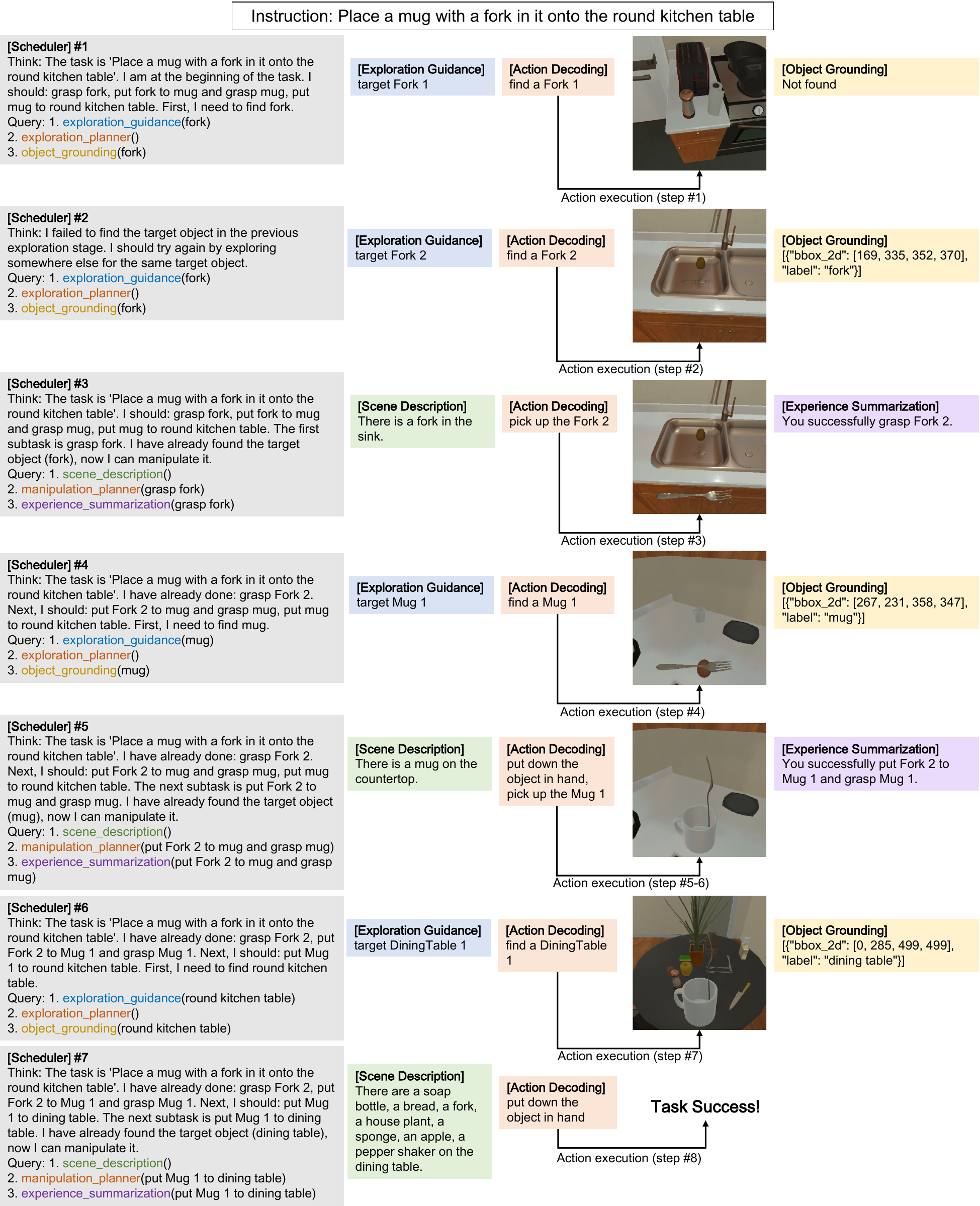}
  \caption{A visualization of the generated plan on EB-ALFRED (visual appearance split).}\label{fig:vis_eb2}
  \vspace{-5mm}
\end{figure*}

\begin{figure*}[h]
  \centering
  \includegraphics[width=0.99\linewidth]{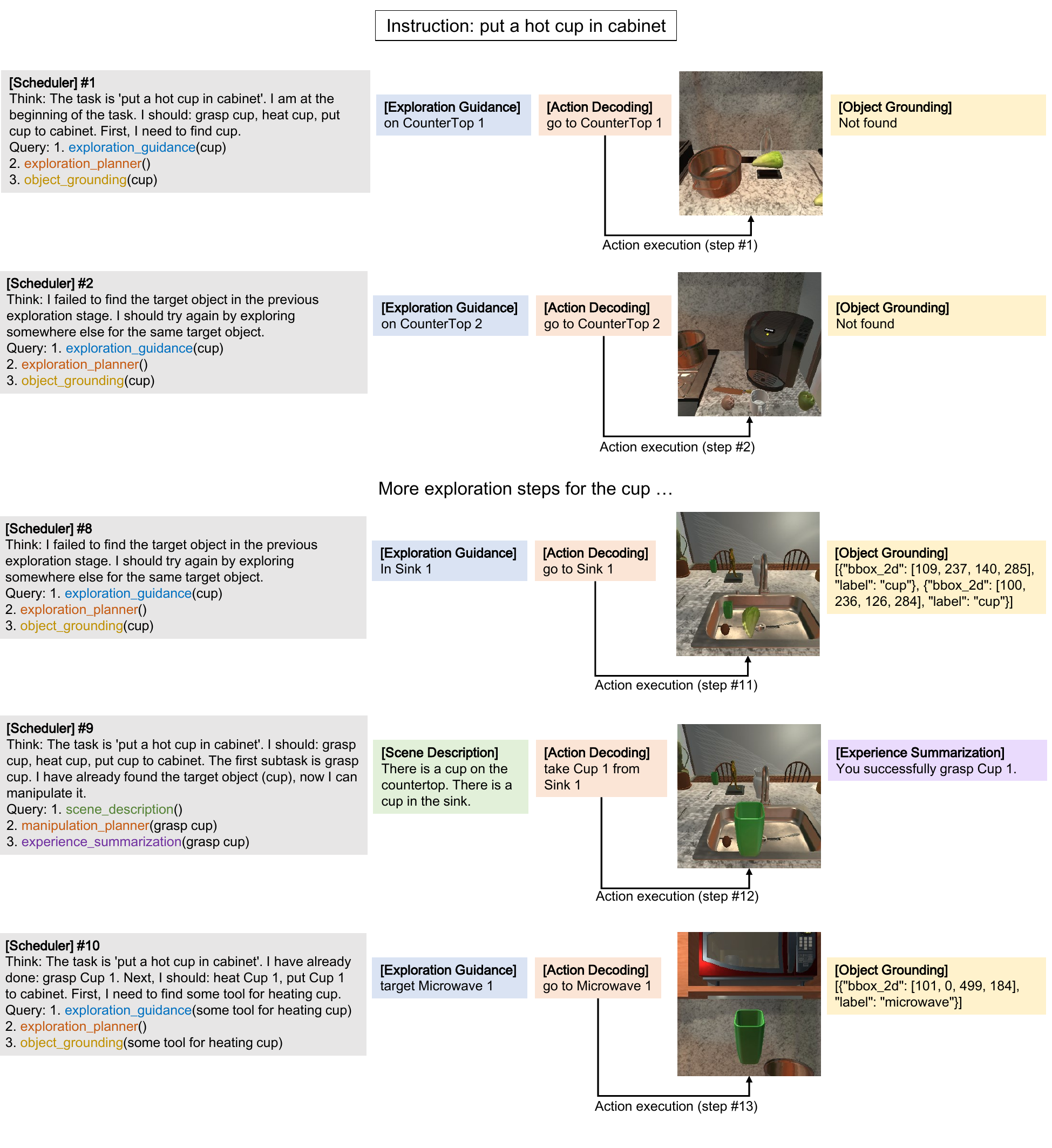}
  \caption{A visualization of the generated plan on ALFWorld's visual environment, part 1.}\label{fig:vis_aw-1}
  \vspace{-5mm}
\end{figure*}

\begin{figure*}[h]
  \centering
  \includegraphics[width=0.99\linewidth]{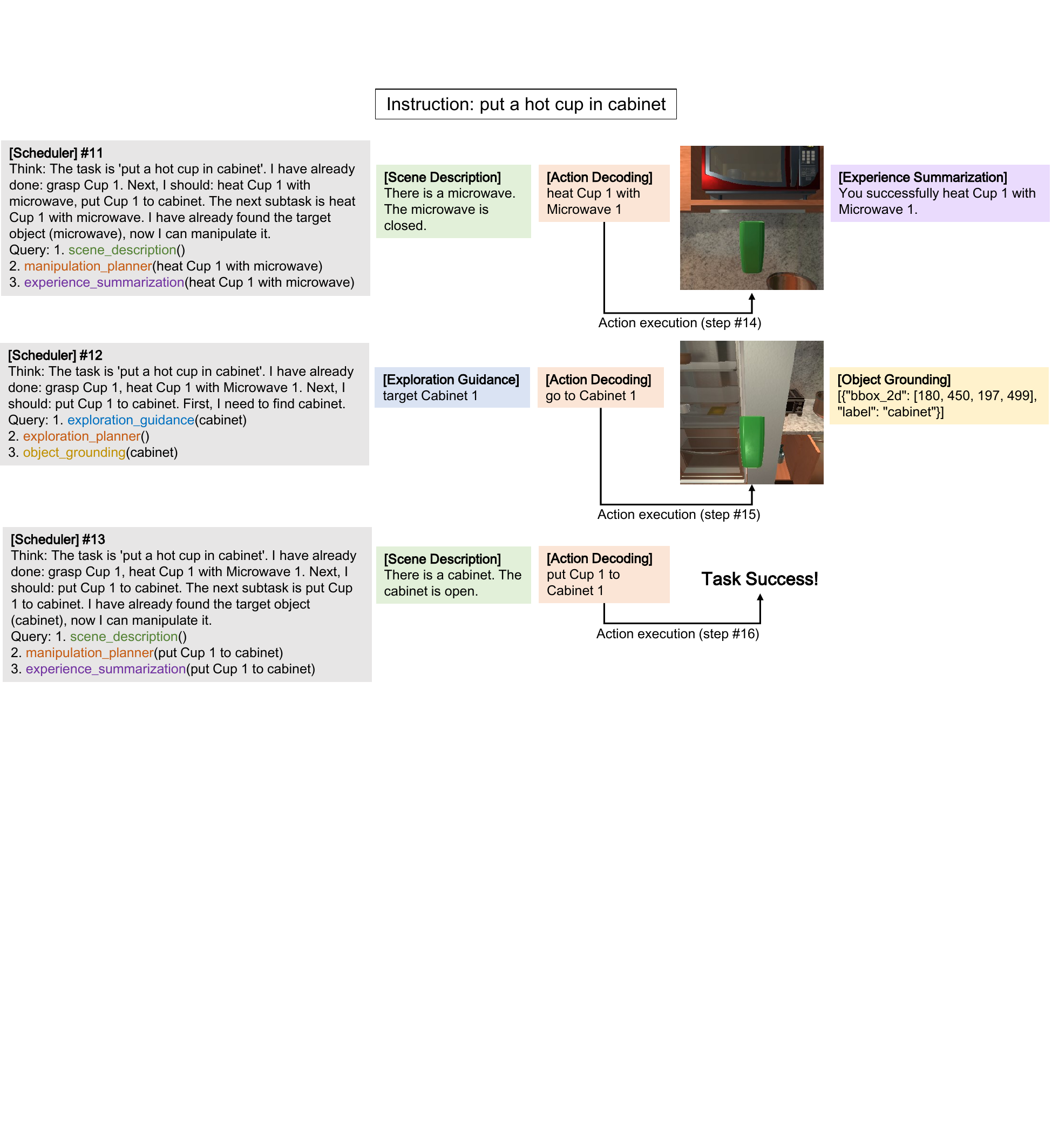}
  \caption{A visualization of the generated plan on ALFWorld's visual environment, part 2.}\label{fig:vis_aw-2}
  \vspace{-5mm}
\end{figure*}

\begin{figure*}[h]
  \centering
  \includegraphics[width=0.99\linewidth]{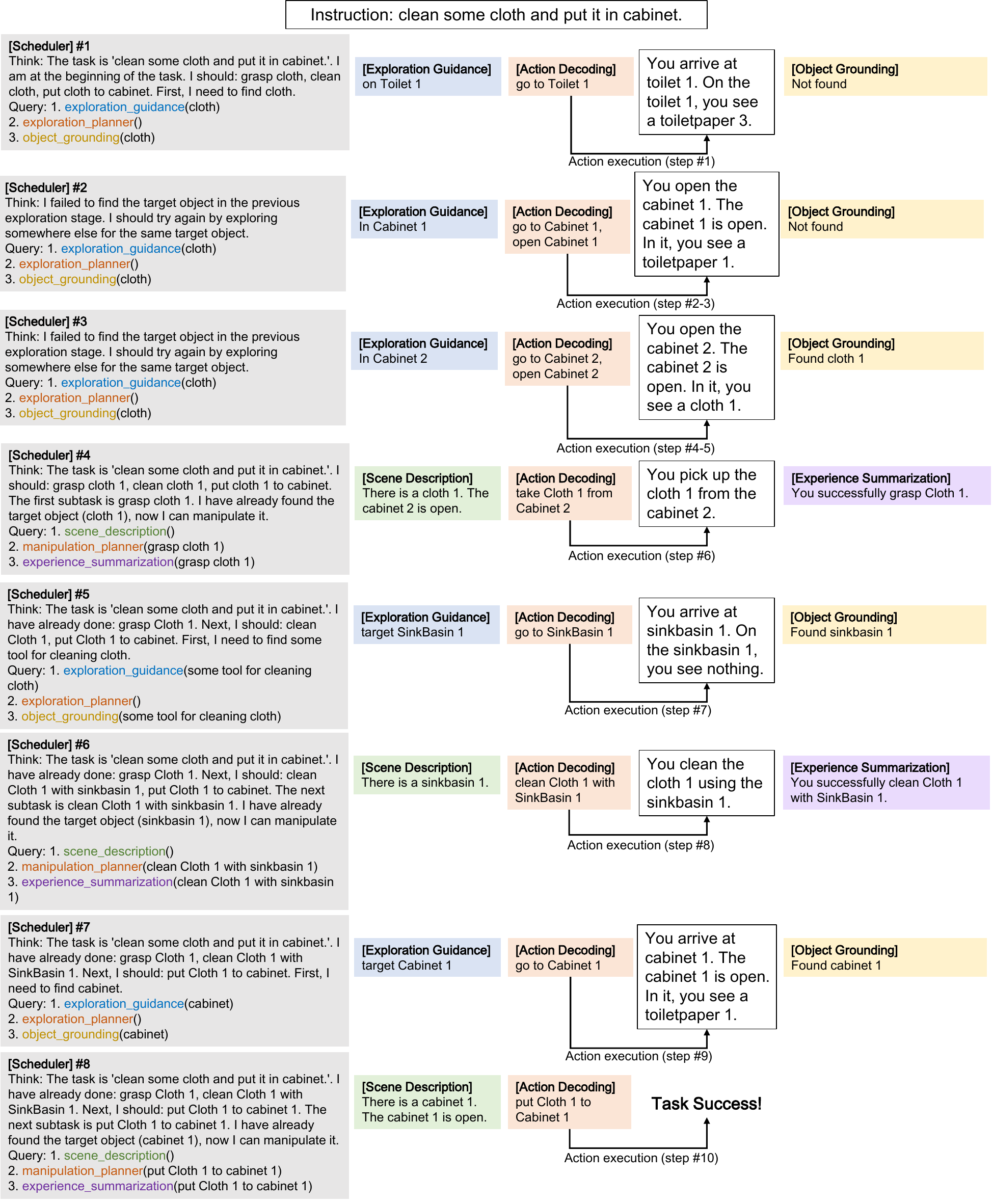}
  \caption{A visualization of the generated plan on ALFWorld's textual environment.}\label{fig:vis_awT}
  \vspace{-5mm}
\end{figure*}

We present the CoT traces and capability invocations generated by RoboAgent on several tasks in Fig.~\ref{fig:vis_eb1-1}-\ref{fig:vis_awT} to qualitatively analyze its planning capabilities. In Fig.~\ref{fig:vis_eb1-1}-\ref{fig:vis_eb1-3}, the agent successfully completes a complex task in EB-ALFRED involving more than 20 action steps, with the scheduler accurately analyzing the task progress and invoking appropriate capabilities to sequentially solve each subtask. Fig.~\ref{fig:vis_eb2} illustrates a task in EB-ALFRED that involves open-vocabulary object reference. The OG capability successfully grounds the query ``round kitchen table” to the dining table, facilitating subsequent planning and manipulation. Fig.~\ref{fig:vis_aw-1}-\ref{fig:vis_aw-2} depict a task in ALFWorld where the agent cannot directly ``go to” small objects. Our EG and OG capabilities assist the agent in efficiently exploring the receptacles in the scene, ultimately locating the target object (cup). Fig.~\ref{fig:vis_awT} shows an example in the text-based environment, where OG, SD, and ES provide feedback by parsing the textual observations.

\subsection{Real-World Demonstration}
\begin{figure*}[h]
  \centering
  \includegraphics[width=0.99\linewidth]{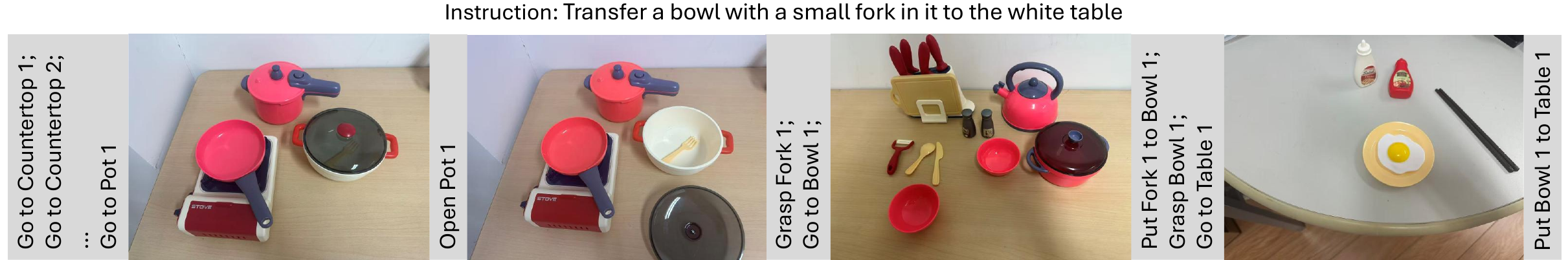}
  \caption{A demonstration of real-world deployment. We present some of the input images and output actions. A human operator carried out the actions and captured the images with camera.}\label{fig:deploy}
\end{figure*}

Following the reviewer’s suggestion, we attempt to deploy RoboAgent in a real-world task setting. To this end, we construct a simple environment using toy kitchen utensils. The scene involves multiple tabletops to simulate the partial observability commonly encountered in household applications. We employ the Qwen3-VL-3B model trained on ALFRED. The task instruction, the list of all receptacles, and manually captured images are provided as inputs to the model, which generates a sequence of actions. The actions are then manually executed by a human operator. The experimental results are illustrated in Fig.~\ref{fig:deploy}. This example provides a preliminary demonstration of the feasibility of RoboAgent when applied to real-world observations.

\section{Error Analysis}\label{sec:EA}
\begin{figure}[h]
  \centering
  \includegraphics[width=0.99\linewidth]{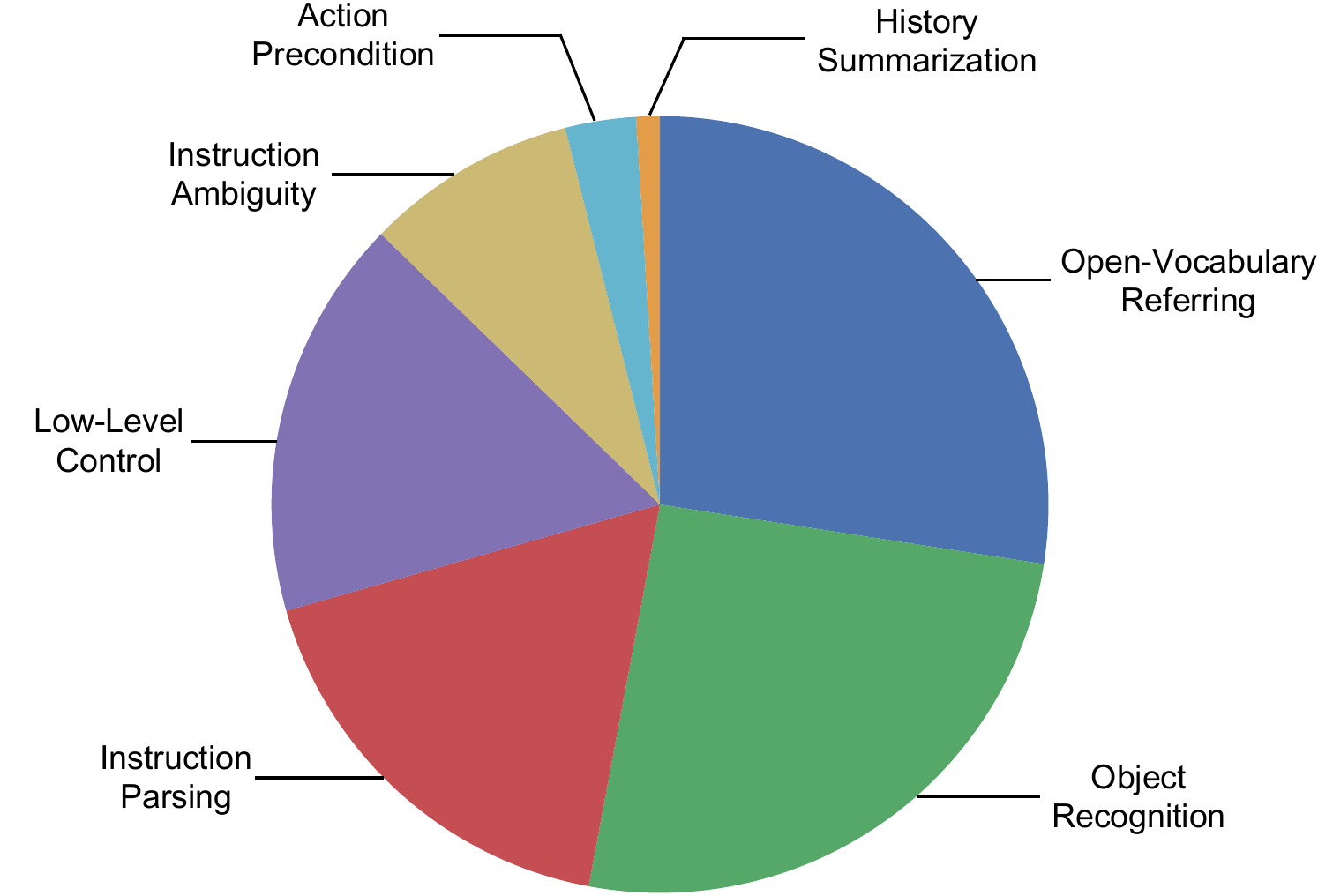}
  \caption{Error distribution on EB-ALFRED.}\label{fig:EA_eb}
\end{figure} 

\begin{figure}[h]
  \centering
  \includegraphics[width=0.99\linewidth]{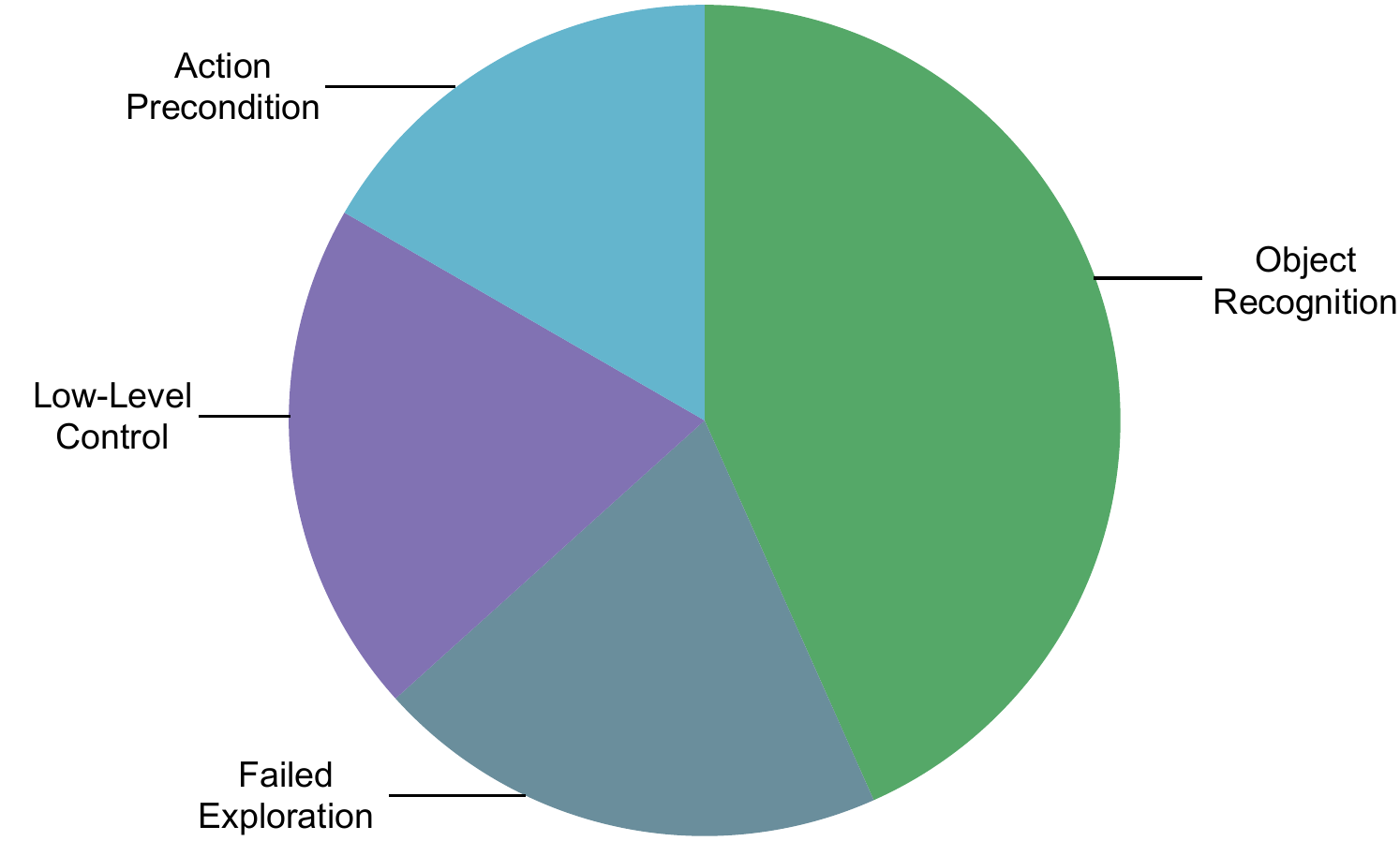}
  \caption{Error distribution on ALFWorld.}\label{fig:EA_aw}
\end{figure} 

\begin{figure}[h]
  \centering
  \includegraphics[width=0.99\linewidth]{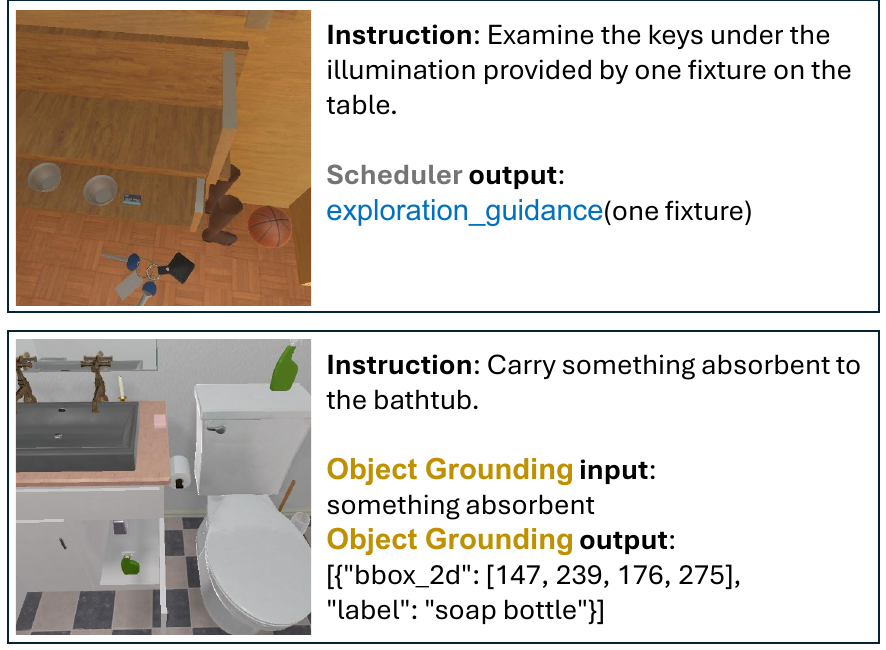}
  \caption{A visualization of two typical failure cases of RoboAgent. In the task above, the scheduler ignores the key information “illumination” from the instruction, and does not provide a complete and correct query to the invoked capabilities. In the task below, OG fails to understand the open-vocabulary query, resulting in the detection of an incorrect target object (soap bottle, instead of towel).}\label{fig:vis_fail}
  \vspace{-5mm}
\end{figure} 

One advantage of our capability-driven task planning framework is that it enables fine-grained failure analysis. We manually examine all failed tasks in EB-ALFRED and ALFWorld and present their attribution in Fig.~\ref{fig:EA_eb} and~\ref{fig:EA_aw}.
For EB-ALFRED, approximately half of the failures come from the visual grounding stage. In some cases, OG fails to recognize the target object in the agent’s field of view (Object Recognition), which may occur when the object is small or occluded. In other cases, EG and OG may not fully understand the open-vocabulary object queries (Open-Vocabulary Referring), causing the agent to miss the target object, interact with a wrong object, or output invalid actions due to wrong object category label. About 18\% of failures are due to the scheduler incorrectly parsing the instruction, which typically occurs when the instruction involves complex sentence structures, irrelevant information, or long object queries. Approximately 17\% of failures originate from low-level control errors, potentially due to simulator implementation issues (\textit{e.g.}, when the agent is near an object but cannot interact with it, or when multiple instances of the same object category appear in the field of view and the agent cannot specify the intended target). Around 9\% of failures result from ambiguities in the instruction itself, such as the agent choosing a soap bottle rather than a spray bottle when instructed to get a ``bottle”, or selecting a dining table instead of a side table for the query ``table.” Another 3\% of errors arise from the ignored action preconditions, \textit{e.g.}, put something to the cabinet when the cabinet is closed, stemming from inaccurate outputs of SD or missing actions in AD. Finally, 1\% of errors are due to the ES capability (History Summarization), where it fails to correctly report the outcome of the previous actions and gives false assumptions for subsequent planning.

In ALFWorld, approximately 43\% of errors also stem from object recognition. Although ALFWorld does not involve open-vocabulary descriptions, the restriction to moving toward receptacles sometimes places the target object at the edge of the observed image, increasing the difficulty of object grounding. About 20\% of failures are due to exploration, where the model exhausts the allowed action steps without finding the target object, indicating room for improvement in EG.\footnote{We also observe that, in some tasks, the target object cannot be seen even when all feasible positions are explored.} Another 20\% of failures are due to low-level control, similarly related to simulator implementation issues. The remaining 17\% arise from ignored action preconditions. We note that, in ALFWorld, the action of ``heat something with microwave" often fails when the microwave is open, suggesting that AD still needs to learn certain simulator-specific action rules.

In Fig.\ref{fig:vis_fail}, we present the visualization of some typical failure cases for the scheduler and the capability.

\end{document}